\documentclass{article}

\usepackage{amsmath,amsfonts,bm}

\def\eqref#1{equation~\ref{#1}}

\def\1{\bm{1}}

\DeclareMathAlphabet{\mathsfit}{\encodingdefault}{\sfdefault}{m}{sl}
\SetMathAlphabet{\mathsfit}{bold}{\encodingdefault}{\sfdefault}{bx}{n}

\usepackage{hyperref}
\usepackage{url}

 \usepackage{ulem}
\usepackage[utf8]{inputenc} 
\usepackage[T1]{fontenc}    
\usepackage{hyperref}       
\usepackage{url}            
\usepackage{booktabs}       
\usepackage{amsfonts}       
\usepackage{nicefrac}       
\usepackage{microtype}      
\usepackage{xcolor}         
\usepackage{color}

\usepackage{graphicx}
\usepackage{subfigure}
\usepackage{kotex}
\usepackage{duckuments}
\usepackage{amsmath}
\usepackage{amssymb}
\usepackage{mathtools}
\usepackage{amsthm}
\usepackage[capitalize,noabbrev]{cleveref}
\usepackage[textsize=tiny]{todonotes}
\usepackage{afterpage}

\usepackage{array}

\usepackage{amsmath,amsfonts,bm}
\usepackage{multirow}
\newcommand{\fref}[1]{Figure~\ref{#1}}

\newcommand{\tref}[1]{Table~\ref{#1}}
\newcommand{\sref}[1]{Section~\ref{#1}}

\newcommand{\onedkd}{Single-attribute Divergence}
\newcommand{\oneabbr}{SaD}
\newcommand{\twodkd}{Paired-attribute Divergence}
\newcommand{\twoabbr}{PaD}
\newcommand{\dcs}{Heterogeneous CLIPScore}
\newcommand{\cs}{CLIPScore}

\def\predictedx0{x_0}

\usepackage{microtype}
\usepackage{graphicx}
\usepackage{subfigure}
\usepackage{booktabs} 

\usepackage{hyperref}

\usepackage[accepted]{icml2024}

\usepackage{amsmath}
\usepackage{amssymb}
\usepackage{mathtools}
\usepackage{amsthm}

\usepackage[capitalize,noabbrev]{cleveref}

\theoremstyle{plain}

\theoremstyle{definition}

\theoremstyle{remark}

\usepackage[textsize=tiny]{todonotes}

\icmltitlerunning{Attribute Based Interpretable Evaluation Metrics for
Generative Models}

\begin{document}

\twocolumn[
\icmltitle{Attribute Based Interpretable Evaluation Metrics for
Generative Models}



\icmlsetsymbol{equal}{*}

\begin{icmlauthorlist}
\icmlauthor{Dongkyun Kim}{equal,yyy,comp}
\icmlauthor{Mingi Kwon}{equal,yyy}
\icmlauthor{Youngjung Uh}{yyy}
\end{icmlauthorlist}

\icmlaffiliation{yyy}{Department of Artificial Intelligence, Yonsei University, Seoul, Republic of Korea}
\icmlaffiliation{comp}{AI Lab, CTO Division, LG Electronics, Seoul, Republic of Korea}

\icmlcorrespondingauthor{Youngjung Uh}{yj.uh@yonsei.ac.kr}

\icmlkeywords{Machine Learning, ICML}

\vskip 0.3in
]



\printAffiliationsAndNotice{\icmlEqualContribution} 

\begin{abstract}
When the training dataset comprises a 1:1 proportion of dogs to cats, a generative model that produces 1:1 dogs and cats better resembles the training species distribution than another model with 3:1 dogs and cats. Can we capture this phenomenon using existing metrics? Unfortunately, we cannot, because these metrics do not provide any interpretability beyond ``diversity". In this context, we propose a new evaluation protocol that measures the divergence of a set of generated images from the training set regarding the \textit{distribution of attribute strengths} as follows. Single-attribute Divergence (SaD) reveals the attributes that are generated excessively or insufficiently by measuring the divergence of PDFs of individual attributes. Paired-attribute Divergence (PaD) reveals such pairs of attributes by measuring the divergence of \textit{joint} PDFs of pairs of attributes. For measuring the attribute strengths of an image, we propose Heterogeneous CLIPScore (HCS) which measures the cosine similarity between image and text vectors with \textit{heterogeneous initial points}. With SaD and PaD, we reveal the following about existing generative models.
ProjectedGAN generates implausible attribute relationships such as \texttt{baby} with \texttt{beard} even though it has competitive scores of existing metrics.
Diffusion models struggle to capture diverse colors in the datasets. The larger sampling timesteps of the latent diffusion model generate the more minor objects including \texttt{earrings} and \texttt{necklace}. Stable Diffusion v1.5 better captures the attributes than v2.1. Our metrics lay a foundation for explainable evaluations of generative models. Code: \href{github.com/notou10/sadpad}{github.com/notou10/sadpad} .

\end{abstract}

\section{Introduction}




The advancement of deep generative models, including VAEs \citep{kingma2013auto}, GANs \citep{karras2019style,karras2020analyzing,karras2021alias,sauer2021projected}, and Diffusion Models (DMs) \citep{song2020denoising,nichol2021improved,rombach2022high}, has led to generated images that are nearly indistinguishable from real ones. Evaluation metrics, especially those assessing fidelity and diversity, play a pivotal role in this progress. One standout metric is Fréchet Inception Distance (FID) \citep{heusel2017gans}, measuring the disparity between training and generated image distributions in embedding space. Coupled with other metrics like precision, recall, density, and coverage, the difference between generated and real image distributions is effectively gauged.

\begin{figure*}[h!]
    \centering
    \includegraphics[width=0.8\textwidth]{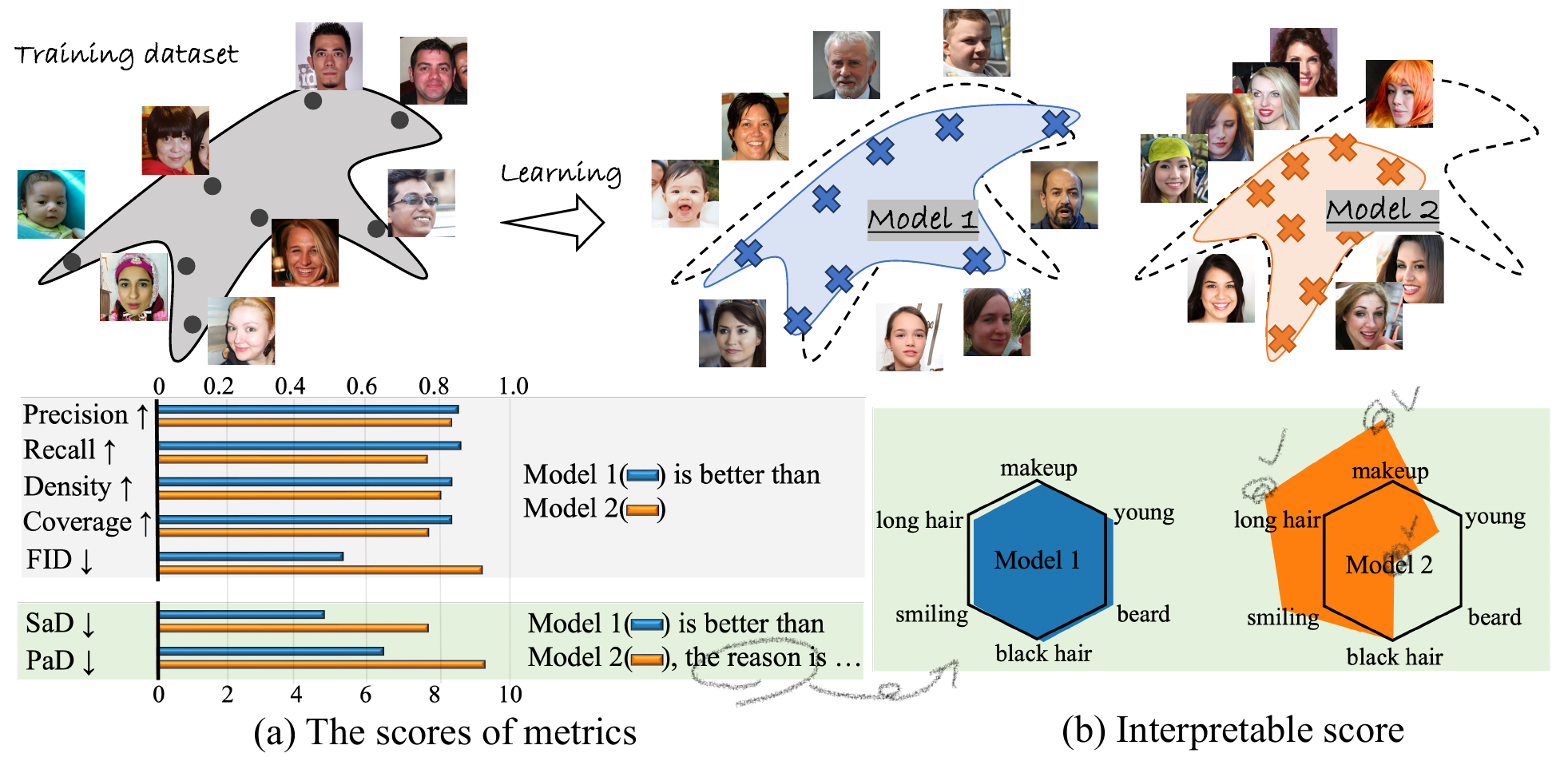}
    \caption{\textbf{Conceptual illustration of our metric.} We design the scenario, Model~2 lacks diversity. (a)~Although existing metrics (gray box) capture the inferiority of Model 2, they do not provide an explanation for the judgments. (b) Our attribute-based proposed metric (green box) has an interpretation: Model~2 is biased regarding \texttt{long hair}, \texttt{makeup}, \texttt{smiling}, and \texttt{beard}.}
    \label{fig:overview}
\end{figure*}

\fref{fig:overview} illustrates the evaluation metrics for two models with distinct properties. While Model~1’s generated images align closely with the training dataset, Model 2 exhibits a lack of diversity. Notably, in \fref{fig:overview}a gray box, Model 1 consistently outperforms Model 2 across all metrics. Yet, these metrics fall short in explicability; for example, they don’t highlight the overrepresentation of \texttt{long hair} and \texttt{makeup} in Model 2.

Addressing this gap, our paper proposes a methodology to quantify discrepancies between generated and training images, focusing on specific attributes. \fref{fig:overview}b shows the concept of our alternative approach that measures the distribution of attribute strengths compared to the training set: while Model 1 offers a balanced attribute distribution akin to the training dataset, Model 2 overemphasizes \texttt{long hair} and underrepresents \texttt{beard}.

To build metrics that quantify the difference between two image sets in an interpretable manner, we introduce  \dcs{} (HCS), an enhanced variant of CLIPScore \citep{radford2021learning}. Compared to CLIPScore, \dcs{} captures the similarity between modalities—image and text—by establishing distinct origins for text and image vectors.

Utilizing HCS, we introduce new evaluation protocols to assess the attribute distribution alignment between generated images and training data as follows.
1) \onedkd{} (\oneabbr{}) measures how much a 
generative model deviates from the distribution of each attribute in the training data. 2) \twodkd{} (\twoabbr{}) measures how much a generative model breaks the relationship between attributes in the training data, such as "babies do not have beards." With the proposed metrics, users can now realize which specific attributes (or pairs of attributes) in generated images differ from those in training images.

\fref{fig:overview}b shows the concept of SaD with 6 attributes, where \texttt{long hair, makeup, beard} are the most influential attributes to SaD. This allows us to explain why Model 2 is not good. We note elaborate quantification of attribute preservation could be one of the meaningful tasks since the generative model can be utilized for diverse purposes such as text-to-image generation not only for generating a plausible image.


We conduct a series of carefully controlled experiments with varying configurations of attributes to validate our metrics in \sref{sec:5.1} and \ref{sec:5.2}. 
Then we provide different characteristics of state-of-the-art generative models \citep{karras2019style,karras2020analyzing,karras2021alias,sauer2021projected,nichol2021improved,rombach2022high,yang2023paint} which could not be seen in the existing metrics.
For instance, GANs better synthesize color-/texture-related attributes such as \texttt{striped fur} which DMs hardly preserve in LSUN-Cat (\sref{sec:5.3}).
When we increase the sampling steps of DMs, tiny objects such as \texttt{necklaces} and \texttt{earrings} tend to appear more frequently.
Even though Stable diffusion v2.1 is reported that have a better FID score than Stable diffusion v1.5, the attribute-aspect score is worse than v1.5 (\sref{sec:stable}).
Our approach is versatile, and applicable wherever image comparisons are needed.
The code will be publicly available.

\section{Related Work}


\paragraph{Fréchet Inception Distance}

Fréchet Inception Distance (FID) \citep{heusel2017gans} calculates the distance between the estimated Gaussian distributions of two datasets using a pre-trained Inception-v3 \citep{szegedy2016rethinking}. However, \citet{kynkaanniemi2022role} pointed out that patterns resembling ImageNet classes significantly influence FID. They proposed using embeddings from the CLIP encoder to make FID less susceptible to intentional or accidental distortions. Additionally, \citet{stein2023exposing} suggested using embeddings from DINO-v2. Despite these suggestions, both approaches merely changed the embedding space for measuring FID, relying on the raw embeddings as they are. In contrast, we design a new representation for this purpose.

\paragraph{Fidelity and diversity}

\citet{sajjadi2018assessing} devised precision and recall for generative model evaluation. Further refinements were provided by \citet{kynkaanniemi2019improved} and \citet{naeem2020reliable}. Generally, these metrics use a pre-trained network to evaluate how embeddings of generated images match with those of real images and vice-versa.



\begin{figure*}[!ht]
    \centering
    \includegraphics[width=0.95\textwidth]{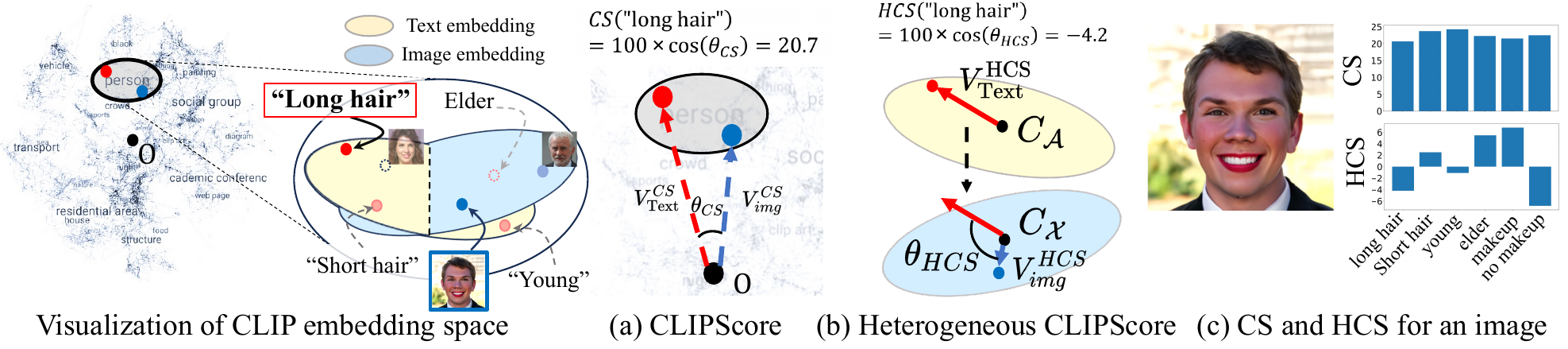}
    \vspace{-0.5em}
     \caption{\textbf{Illustration of CLIPScore and \dcs{}.} We visualized the CLIP embedding space obtained from multiple texts. The yellow ellipse represents the embedding space of CelebA’s text attributes, while the blue ellipse visualizes the embedding space of images. (a) CLIPScore (CS) evaluates the similarity between $V^{CS}_{img}$ and $V^{CS}_{Text}$ from the coordinate origin, where the angle between the two vectors is bounded, resulting in a limited similarity value. (b)  \dcs{} (HCS) gauges the similarity between $V^{HCS}_{img}$ and $V^{HCS}_{Text}$ using the defined means of images $C_\mathcal{X}$ and texts $C_\mathcal{A}$ as the origin, the range of similarity is unrestricted. (c) shows flexible values of HCS compared to CS. } 
    \vspace{-1em}
     \label{fig:2}
\end{figure*}

\paragraph{Other metrics}


Beyond these, metrics such as  Kernel Inception Distance (KID) \citep{binkowski2018demystifying}, Perceptual path length \citep{karras2019style}, Fréchet segmentation distance \citep{bau2019seeing}, and Rarity score \citep{han2022rarity} have been introduced. The first calculates squared Maximum Mean Discrepancy (MMD) between inception representations, the second indicates latent space smoothness, the third measures pixel segmentation differences, and the latter assesses the rarity of generated images.
However, these metrics predominantly rely on raw embeddings from pretrained classifiers, yielding scores with limited interpretability. As \fref{fig:overview}a indicates, while some metrics highlight poor image generation performance, they lack in-depth explanatory insights. We aim to fill this gap with our novel, detailed, and insightful evaluation metrics.

TIFA \citep{hu2023tifa} uses visual question answering to validate if text-to-image results correspond to the input texts. On the other hand, our metrics evaluate the distribution of attribute strengths in a set of images.

\section{Toward Explainable Metrics}


Existing metrics for evaluating generated images often use embeddings from Inception-V3 \citep{szegedy2016rethinking} or CLIP image encoder \citep{dosovitskiy2020image}. Yet, these embeddings lack clarity in interpreting each channel in the embedding. Instead, we opt to measure attribute strengths in images for a predefined set of attributes. We first explain CLIPScore as our starting point (\sref{sec:embedding}), introduce Heterogeneous CLIPScore (\sref{sec:dcs}), and describe ways of specifying the target attributes (\sref{sec:selection}.)


\subsection{Measuring attribute strengths with CLIP}
\label{sec:embedding}


For a set of attributes, we start by measuring the attribute strengths of images. The typical approach is computing CLIPScore:
\begin{equation}
\label{eq1}
\text{CLIPScore}(x,a) = 100 \times \text{sim}(\mathbf{E}_{\mathbf{I}}(x),\mathbf{E}_{\mathbf{T}}(a)),
\end{equation}
where $x$ is an image, $a$ is a given text of attribute, $\text{sim}(*,*)$ is cosine similarity, and $\mathbf{E}_{\mathbf{I}}$ and $\mathbf{E}_{\mathbf{T}}$ are CLIP image encoder and text encoder, respectively.
\fref{fig:2}c shows an example CLIPScores of an image regarding a set of attributes. Yet, CLIPScores themselves do not provide a clear notion of attribute strengths as we observe ambiguous similarities between opposite attributes. The research community is already aware of such a problem. 
To overcome this, we introduce \dcs{} in the subsequent subsections, showcased in \fref{fig:2}c, ensuring more accurate attribute strengths.

\subsection{\dcs{}}
\label{sec:dcs}

In the earlier section, we noted that CLIPScore tends to have a narrow value range, as visualized in \fref{fig:2}a. To remedy this, we introduce \dcs{} (HCS). It uses heterogeneous initial points for image and text embedding vectors as follows.



Given training images denoted as $\{x_1, x_2, ..., x_{\textit{N}_{\mathcal{X}}}\} \in \mathcal{X}$, and a set of attributes defined as $\{a_1, a_2, ..., a_{\textit{N}_{\mathcal{A}}}\} \in \mathcal{A}$, we define $C_\mathcal{X}$ as the center of images and $C_\mathcal{A}$ as another center of text attributes on CLIP embedding, respectively as

\begin{equation}\label{eq2}
C_\mathcal{X} = \frac {1}{\textit{N}_{\mathcal{X}}}\sum_{i=1}^{\textit{N}_{\mathcal{X}}}\textbf{E}_{\textbf{I}}(x_i), \quad
C_{\mathcal{A}} = \frac {1}{\textit{N}_{\mathcal{A}}}\sum_{i=1}^{\textit{N}_{\mathcal{A}}}\textbf{E}_{\textbf{T}}(a_i).
\end{equation}


These centers act as initial points of the embedding vectors. HCS is defined by the similarity between the two vectors, $V_x$ and $V_{a}$. The former connects the image center to a specific image, while the latter connects the attribute center to a particular attribute. Then we define 

\begin{equation}\label{eq3}
V_{x} = \textbf{E}_{\textbf{I}}(x) - C_\mathcal{X}, \quad
V_{a} = \textbf{E}_{\textbf{T}}(a) - C_\mathcal{A},
\end{equation}
\begin{equation}
\label{eq4}
\text{HCS}(x,a) = 100 \times \text{sim}(V_{x}, V_{a}),
\end{equation}

where $\text{sim}(*,*)$ computes cosine similarity. For extending HCS from a single sample to all samples, we denote the probability density function (PDF) of $\text{HCS}(x_i, a_i)$ for all $x_i \in \mathcal{X}$ as $\text{HCS}_\mathcal{X}(a_i)$.

\fref{fig:2} illustrates the difference between HCS (\dcs{}) and CS (CLIPScore). HCS uses the respective centers as initial points, allowing for clearer determination of attribute magnitudes, whereas CS lacks this clarity.




\subsection{Attribute selection}
\label{sec:selection}

The effectiveness of our evaluation metric is contingent upon the target attributes we opt to measure. Inspired by recent work \citep{hu2023tifa} evaluate generative models utilizing large language model (LLM), we employ a vision-language model (VLM) to determine the best attributes that truly capture generator performance. We pinpoint and assess the attributes evident in the training data via image descriptions. By analyzing the frequency of these attributes in image captions, we can identify which ones are most prevalent. To achieve this for captionless datasets, we employ the image captioning model, BLIP \citep{li2022blip}, to extract words related to attributes from the training data. We then adopt $N$ frequently mentioned ones as our target attributes, denoted as $\mathcal{A}$, for the metric. Given that these attributes are derived automatically, utilizing BLIP for this extraction could serve as a foundational method.
In our \tref{tab:allignment}, we demonstrate that using BLIP to extract attributes exhibits similar tendencies to both using CelebA GT labels and defining attributes through LLM.

\section{Evaluation Metrics with Attribute Strengths}
In this section, we harness the understanding of attribute strengths to devise two comprehensible metrics. Section \ref{subsec:sakd} introduces \onedkd{} (\oneabbr{}), quantifying the discrepancy in attribute distributions between training data and generated images. Section \ref{subsec:pakd} brings forth \twodkd{} (\twoabbr{}), evaluating the relationship between attribute strengths.



\subsection{Single-attribute Divergence}
\label{subsec:sakd}

If we have a dataset with dogs and cats, and a generative model only makes dog images, it is not an ideal model because it does not produce cats at all \citep{goodfellow2016deep}. With this idea, we say one generative model is better than another if it makes a balanced number of images for each attribute similar to the training dataset. Since we do not know the true distribution of real and fake images, we came up with a new metric, Single-attribute Divergence (SaD). This metric checks how much of each attribute is in the dataset by utilizing interpretable representation. Our metric, \oneabbr{}, quantifies the difference in density for each attribute between the training dataset ($\mathcal{X}$) and the set of generated images ($\mathcal{Y}$). We define \oneabbr{} as
\begin{equation}\label{eq5}
\text{SaD}(\mathcal{X},\mathcal{Y}) = \frac {1}{\textit{M}} \sum_{i}^{\textit{M}} \text{KL}(\text{HCS}_{\mathcal{X}}(a_{i}), \text{HCS}_{\mathcal{Y}}(a_{i})),
\end{equation}

where $i$ denotes an index for each attribute, \textit{M} is the number of attributes, KL(*) is Kullback-Leibler divergence, and $\text{HCS}_\mathcal{X}(a_i)$ denotes PDF of $\text{HCS}(x_i, a_i)$ for all $x_i \in \mathcal{X}$.

We first estimate PDFs of \dcs{} for each attribute present in $\mathcal{X}$ and $\mathcal{Y}$  by applying Gaussian Kernel Density Estimation (KDE) on the entire sample for each dataset. Subsequently, we compare these HCS PDFs which reflect the distribution of attribute strengths within the datasets. If an attribute's distribution in $\mathcal{X}$ closely mirrors that in $\mathcal{Y}$, their respective HCS distributions will align, leading to similar PDFs. To measure discrepancies between these distributions, we employ Kullback-Leibler Divergence (KLD). This quantifies how much the generated images either over-represent or under-represent specific attributes compared to the original data.
Subsequently, we determine the average divergence across all attributes between $\mathcal{X}$ and $\mathcal{Y}$ to derive the aggregated metric for \oneabbr{}.


In addition, we define the mean difference of attribute strength to further examine whether poor SaD comes from excessive or insufficient strength of an attribute $a$:
\begin{equation}\label{eq7}
\abovedisplayskip=0pt
\text{mean difference} =\frac {1}{\textit{N}_x} \sum_{i}^{\textit{N}_x}\text{HCS}(x_i,a) - \frac {1}{\textit{N}_y} \sum_{i}^{\textit{N}_y}\text{HCS}(y_i,a).
\end{equation}

where $N_x$ and $N_y$ are the number of training images and generated images, respectively.
Intuitively, a high magnitude of mean difference indicates the mean strength of $\mathcal{Y}$ differs significantly from $\mathcal{X}$ for attribute $a$. A positive value indicates $\mathcal{Y}$ has images with stronger $a$ than $\mathcal{X}$, and vice versa for a negative value. While this does not conclusively reveal the exact trend due to $a$'s complex distribution, it provides an intuitive benchmark.

\subsection{Paired-attribute Divergence}
\label{subsec:pakd}

We introduce another metric, Paired-attribute Divergence (\twoabbr{}), aimed at evaluating whether generated images maintain the inter-attribute relationships observed in the training data. Essentially, if specific attribute combinations consistently appear in the training data, generated images should also reflect these combinations. To illustrate, if every male image in the training dataset is depicted wearing glasses, the generated images should similarly represent males with glasses. We assess this by examining the divergence in the joint probability density distribution of attribute pairs between the training data and generated images. This metric, termed Paired-attribute Divergence (\twoabbr{}), leverages joint probability density functions as detailed below:
\begin{equation}\label{eq6}
\text{PaD}(\mathcal{X},\mathcal{Y}) = \frac {1}{\left| \textit{P} \right|} \sum_{(i,j)}^{\textit{P}} \text{KL}(\text{HCS}_{\mathcal{X}}(a_{i,j}), \text{HCS}_{\mathcal{Y}}(a_{i,j})),
\end{equation}

where $M$ is the number of attributes, $P = \binom{M}{2}$, $(i,j)$ denotes an index pair of attributes selected out of $M$, and the joint PDF of the pair of attributes is denoted as $\text{HCS}_{\mathcal{X}}(a_{i,j})$.


When utilized together with SaD, \twoabbr{} will offer a comprehensive analysis of the model's performance. For instance, if the probability density function of the generator for the attribute pair (\texttt{baby}, \texttt{beard}) diverges notably from the training data's distribution while SaD for \texttt{baby} and \texttt{beard} are comparatively low, it suggests that the generator may not be effectively preserving the (\texttt{baby}, \texttt{beard}) relationship. Consequently, \twoabbr{} enables us to quantify how well attribute relationships are maintained in generated data. To the best of our knowledge, we are the first to propose a metric for interdependencies relationship. Moreover, this becomes interpretable.

\section{Experiments}

\paragraph{Experiment details}
For estimating the probability density function (PDF) of \dcs{} (HCS) in both the training data and generated images, we use Gaussian Kernel Density Estimation (KDE). In this process, we extract 10,000 samples from generated and real images to obtain PDFs of attribute strengths. These PDFs are then used to compute \oneabbr{} and \twoabbr{}. In every experiment, we use a set of $\textit{N}_{\mathcal{A}}=20$ attributes. In the case of the toy experiments on FFHQ, we use attributes from CelebA ground truth label (\tref{tab:attr_list}) for the convenience of interpretation.






\subsection{\dcs{} outperforms \cs{}}
\label{sec:5.0}

Heterogeneous CLIPScore (HCS) outshines CLIPScore (CS) in binary classification of the attributes in CelebA. \tref{tab:CS_DCS2} reports accuracy and F1 score computed as follows. For each attribute, we sort the scores of test images and classify the images with top $k$ scores as positives to compute accuracy and F1 score where $k$ denotes the number of positive images in the test set. Then we compute their mean over all attributes. This superiority persists even for refined attributes which excludes subjective attributes such as $\texttt{attractive}$ or $\texttt{blurry}$ as shown in \tref{tab:DSC_from_img_mean}. \tref{tab:celebA_refine_attributes} provides the full list of attributes and the refined attributes. More details are available in the Appendix \ref{sec:accuracy}.


\begin{table}[!t]
\centering
\caption{\textbf{CLIPScore and \dcs{}'s  accuracy on CelebA dataset.}}
\begin{tabular}{c|cc}
                       & accuracy       & f1 score \\ \hline
\dcs{} & \textbf{0.817} & \textbf{0.616}    \\
CLIPScore                     & 0.798 & 0.575   
\end{tabular}
\label{tab:CS_DCS2}
\end{table}

\subsection{Biased data injection experiment: the effectiveness of our metric}
\label{sec:5.1}



\begin{figure}[!t]
\centering
\includegraphics[width=1\linewidth]{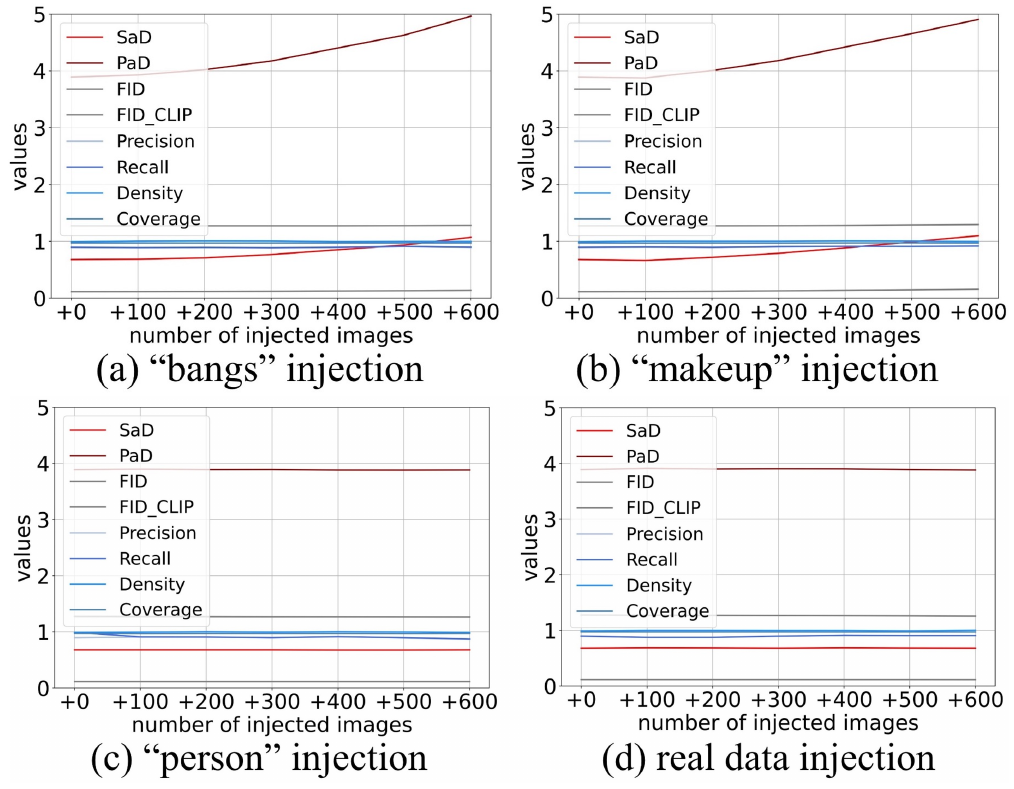}
\caption{\textbf{Validation of metrics through biased injection.} We design one set: typical 30K of FFHQ images, and another set: 30K FFHQ + injected images. Biased data injection, illustrated in (a) with \texttt{makeup} and (b) with \texttt{bangs} leads to an increase in both SaD and PaD rise. In contrast, unbiased data injection (c) \texttt{person} and (d) real data, injecting the same distribution as the training set results in no SaD and PaD rise. Our metrics effectively capture changes in attribute distribution, while existing metrics cannot.}
\vspace{-0em}
\label{fig:injection}
\end{figure}

In this subsection, we conduct a toy experiment to validate our metrics against existing methods. Initially, two non-overlapping subsets, each with 30K images from FFHQ, are defined as training data $\mathcal{X}$ and generated images $\mathcal{Y}$. Starting with these subsets that share a similar distribution, we gradually infuse biased data into $\mathcal{Y}$. 
The biased data is generated using DiffuseIT \citep{kwon2022diffusIT}. We translate samples from the training data, without overlap to the initial 60K images, into \texttt{makeup} (\fref{fig:injection}a) and \texttt{bangs} (\fref{fig:injection}b). We also provide controlled counterpart where injected samples are unbiased data translated into the \texttt{person} (\fref{fig:injection}c), or injected samples remain untranslated  
(\fref{fig:injection}d).

As depicted in \fref{fig:injection}, our metrics display a consistent trend: SaD and PaD rise with the inclusion of more edited images in $\mathcal{Y}$, whereas other metrics are static. Thanks to the attribute-based design, our metric suggests that \texttt{makeup} or \texttt{bangs} is the dominant factor for SaD, and relationships that are rarely seen in training data such as (\texttt{man}, \texttt{makeup}) and (\texttt{man}, \texttt{bangs}) for PaD. The impact on SaD and PaD scales linearly with 
the number of images from different attribute distributions.
For an expanded discussion and additional experiments, refer to \fref{fig:injection_appendix} and Appendix \ref{subsec:skip}. These results underscore that \oneabbr{} adeptly discerns the attribute distribution variation, and \twoabbr{} identifies the joint distribution shift between attribute pairs, outperforming other metrics.

\subsection{Discernment of \twoabbr{}}
\label{sec:5.2}
In another toy experiment, we designed a scenario where \oneabbr{} metric struggled to detect specific attribute relationships, while \twoabbr{} metric successfully pinpointed them. We curate three sets of images sampled from CelebA, with the identical marginal distribution of individual attributes. \textbf{set~A}: randomly sampled with a restriction: only \texttt{men} wear \texttt{eyeglasses}, \textbf{set~B}: randomly sampled with a restriction: only \texttt{men} wear \texttt{eyeglasses} (identical to set~A), and \textbf{set~C}: randomly sampled with a restriction: only \texttt{women} wear \texttt{eyeglasses} (corrupted correlation). 

We then observe whether SaD and PaD capture difference due to the corrupted correlation between gender (\texttt{men} or \texttt{women}) and \texttt{eyeglasses}, measuring SaD and PaD of \textbf{set B} and \textbf{set C} from \textbf{set A}. \tref{tab:discern} shows PaD correctly reveals the most influential pair of attributes (\texttt{men} \& \texttt{eyeglasses}) by 26 times higher PaD than the mean PaD, while SaD struggles to capture corruption of correlation between gender (\texttt{men} or \texttt{women}) and \texttt{eyeglasses}. This highlights the effectiveness of PaD in capturing errors in pairwise relations, and the necessity of employing PaD for comprehensive model analysis. Full SaD and PaD is available in \fref{fig:eyeglass_100}.


\begin{table}[]
\caption{\textbf{Discernment of PaD.} \tref{tab:discern} shows SaD to be consistent, while PaD to be different because the correlation between gender and eyeglasses is corrupted. We use 10,000 images for each set, and marginals (the number of images for \texttt{men}, \texttt{women} and \texttt{eyeglasses}) are the same across set A, B, and C.}
\centering
\resizebox{\linewidth}{!}{
\begin{tabular}{c|ccc}
                     & SaD  & PaD      & PaD(\texttt{men} \& \texttt{eyeglasses}) \\ \hline
between set A \& set B & 1.52 & 6.14     & 5.68              \\
between set A \& set C & 1.51 & \textbf{9.60} & \textbf{251.34}         
\end{tabular}}
\label{tab:discern}
\end{table}

\begin{table*}[!t]
\centering
\caption{\textbf{Comparing the performance of generative models.} We computed each generative model's performance on our metric with their official pretrained checkpoints on FFHQ \citep{karras2019style}. We used 50,000 images for both GT and the generated set.} 
\centering
\resizebox{\textwidth}{!}{
\begin{tabular}{c|cccccccc}
                     & StyleGAN1 & StyleGAN2      & StyleGAN3 & iDDPM         & LDM (50) & LDM (200)   & StyleSwin & ProjectedGAN  \\ \hline
SaD ($10^{-7}$)\textdownarrow                 & 11.35     & \textbf{7.52}  & 7.79      & 14.78     & 10.42    & 14.04 & 10.76     & 17.61         \\
PaD ($10^{-7}$)\textdownarrow                 & 27.25     & \textbf{19.22} & 19.73     & 34.04      & 25.36   & 30.71 & 26.56     & 41.53         \\ \hline
FID\textdownarrow                   & 4.74      & \textbf{3.17}  & 3.20      & 7.31      & 12.18    & 11.86 & 4.45      & 5.45          \\ 
FID$\mathrm{_{CLIP}}$\textdownarrow             & 3.17      & \textbf{1.47}  & 1.66      & 2.39   & 3.89       & 3.57  & 2.45      & 3.63          \\
Precision\textuparrow            & 0.90      & 0.92           & 0.92      &0.93 &  \textbf{0.94} & 0.91  & 0.92      & 0.92          \\
Recall\textuparrow               & 0.86      & 0.89           & 0.90      & 0.84        & 0.82  & 0.88  & 0.91      & \textbf{0.92} \\
Density\textuparrow              & 1.05      & 1.03           & 1.03      & \textbf{1.09}  & 1.09 & 1.07 & 1.01      & 1.05          \\
Coverage\textuparrow             & 0.97      & 0.97           & 0.97      & 0.95         & 0.94  & 0.97 & 0.97      & 0.97         

\end{tabular}}
\label{tab:gan_diff_blip}
\end{table*}


\subsection{Comparing generative models with our metrics}
\label{sec:5.3}



Leveraging the superior sensitivity and discernment of our proposed metrics, we evaluate the performance of GANs and Diffusion Models (DMs) in Table \ref{tab:gan_diff_blip}. Generally, the tendency of SaD and PaD align with other existing metrics. However three notable points emerge; 1) ProjectedGAN \citep{sauer2021projected} lags in performance, 2) As sampling timesteps in DM increase, FIDs improve, while SaD and PaD decline. 3) GANs and Diffusion models vary in their strengths and weaknesses concerning specific attributes.



\begin{figure}[!t]
\centering
\includegraphics[width=0.6\linewidth]{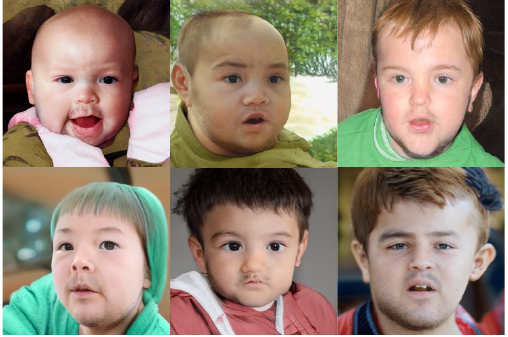}
\caption{\textbf{Failure cases by ProjectedGAN.} ProjectedGAN disregards attribute relationships, such as generating babies with beards.}
\vspace{-0em}
\label{fig:pjgan}
\end{figure}

1) ProjectedGAN~\citep{sauer2021projected} prioritizes matching the training set’s embedding statistics for improving FID rather than improving actual fidelity~\citep{kynkaanniemi2022role}. While it performs well in existing metrics, it notably underperforms in SaD and particularly in PaD. This implies that directly mimicking the training set’s embedding stats does not necessarily imply correct attribute correlations. \fref{fig:pjgan} provides failure cases generated by ProjectedGAN. 

\begin{figure}[!t]
    \centering
    \includegraphics[width=0.8\linewidth]{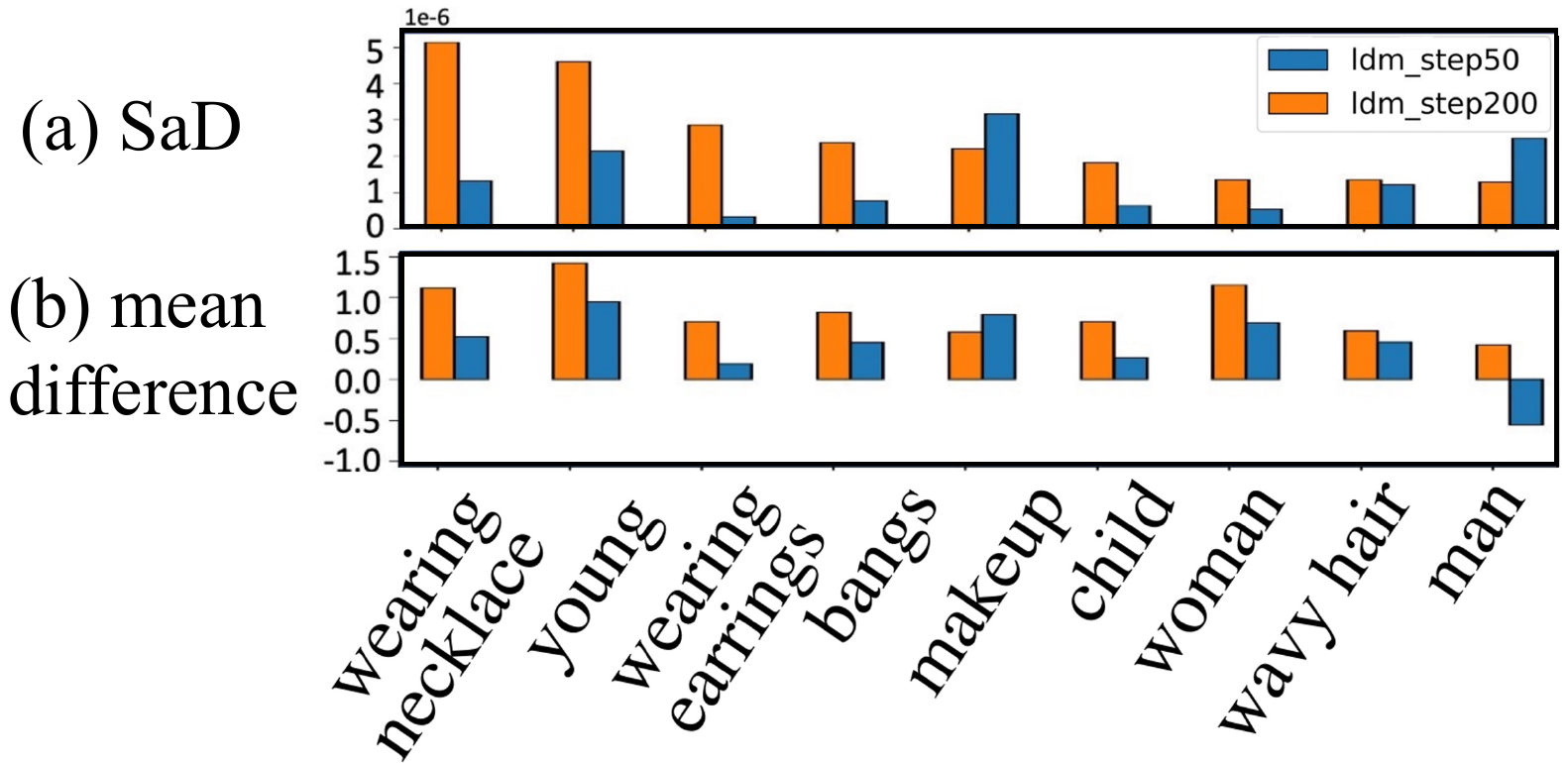}
    \caption{\textbf{LDM with 50 steps v.s. LDM with 200 timesteps.} With increased sampling timesteps, (a) SaD of LDM gets worse, (b) since making too many fine objects such as \texttt{earrings} or \texttt{necklace}.} 
    \vspace{-0.5em}
     \label{fig:5}
\end{figure}

2) Diffusion models typically yield better quality with higher number of sampling timesteps. Yet, SaD and PaD scores for LDM with 200 steps surpass those of LDM with 50 steps. As illustrated in \fref{fig:5}, higher sampling timesteps in the LDM model produce more high-frequency elements such as \texttt{necklaces} and \texttt{earrings}. This could explain the dominance of attributes such as \texttt{young, makeup, woman, wavy hair} naturally. We suppose that dense sampling trajectory generates more high-frequency objects. The scores and mean differences of each attribute are depicted in \fref{fig:5}a and \fref{fig:5}b respectively.

3) Diffusion models fall short on modeling color-related attributes than shape-related attributes. As our metrics provide flexible customization, we report SaD and PaD of color attributes (e.g., \texttt{yellow fur}, \texttt{black fur}) and shape attributes (e.g., \texttt{pointy ears}, \texttt{long tail}) within LSUN Cat dataset. \tref{tab:color_shape} shows that iDDPM excels in matching shape attributes compared to color attributes.

This aligns with the hypothesis by \citet{khrulkov2022understanding} suggesting that DMs learn the Monge optimal transport map, the shortest trajectory, from Gaussian noise distribution to image distribution regardless of training data. This implies that when the initial latent noise $x_T$ is determined, the image color is also roughly determined because the diffused trajectory tends to align with the optimal transport map.

In addition, iDDPM shows notable scores, with the attribute \texttt{arched eyebrows} showing scores over two times higher than GANs in SaD, and attributes related to \texttt{makeup} consistently receive high scores across all StyleGAN 1, 2, and 3 models in PaD. Investigating how the generation process of GANs or DMs affects attributes such as attributes would be an intriguing avenue for future research. See Appendix \ref{sec:appen_analysis} for details.
\begin{table}[]
\centering
\caption{\textbf{SaD and PaD of models with different attributes for LSUN Cat.} Analyzing the weakness of iDDPM for specific attribute types, such as color or shape.}
\label{tab:lsun_cat}
\resizebox{\columnwidth}{!}{%
\begin{tabular}{c|cc|cc}
                                                                                                 & \multicolumn{2}{c|}{color attributes}                                                                                                             & \multicolumn{2}{c}{shape attributes}                                                                                                              \\
                                                                                                 & \begin{tabular}[c]{@{}c@{}}SaD\\ ($10^{-7}$)\textdownarrow\end{tabular} & \begin{tabular}[c]{@{}c@{}}PaD\\ ($10^{-7}$)\textdownarrow\end{tabular} & \begin{tabular}[c]{@{}c@{}}SaD\\ ($10^{-7}$)\textdownarrow\end{tabular} & \begin{tabular}[c]{@{}c@{}}PaD\\ ($10^{-7}$)\textdownarrow\end{tabular} \\ \hline
\multirow{2}{*}{\begin{tabular}[c]{@{}c@{}}StyleGAN1\\ \citep{karras2019style}\end{tabular}}     & \multirow{2}{*}{\textbf{139.03}}                                        & \multirow{2}{*}{\textbf{248.96}}                                        & \multirow{2}{*}{169.76}                                                 & \multirow{2}{*}{318.46}                                                 \\
                                                                                                 &                                                                         &                                                                         &                                                                         &                                                                         \\
\multirow{2}{*}{\begin{tabular}[c]{@{}c@{}}StyleGAN2\\ \citep{karras2020analyzing}\end{tabular}} & \multirow{2}{*}{\textbf{112.06}}                                        & \multirow{2}{*}{\textbf{195.75}}                                        & \multirow{2}{*}{132.41}                                                 & \multirow{2}{*}{246.44}                                                 \\
                                                                                                 &                                                                         &                                                                         &                                                                         &                                                                         \\ \hline
\multirow{2}{*}{\begin{tabular}[c]{@{}c@{}}iDDPM\\ \citep{nichol2021improved}\end{tabular}}      & \multirow{2}{*}{46.93}                                                  & \multirow{2}{*}{85.99}                                                  & \multirow{2}{*}{\textbf{32.48}}                                         & \multirow{2}{*}{\textbf{62.69}}                                         \\
                                                                                                 &                                                                         &                                                                         &                                                                         &                                                                        
\end{tabular}}
\label{tab:color_shape}
\vspace{-1em}
\end{table}


\subsection{Evaluating text-to-image models}
\label{sec:stable}
Recently, there has been a huge evolution of text-to-image generative models \citep{nichol2021glide,rombach2022high,saharia2022photorealistic,balaji2022ediffi}. To evaluate text-to-image models, zero-shot FID score on COCO \citep{lin2014microsoft} is widely used including Stable Diffusion (SD).
Instead, we use our metrics to examine text-to-image models regarding excessively or insufficiently generated attributes.
We generate 30K images with captions from COCO using SDv1.5 and SDv2.1 to calculate SaD and PaD with attributes extracted from the captions. We use $N_{\mathcal{A}}=30$.

\tref{tab:stable} shows SDv1.5 has twice better SaD and PaD than SDv2.1. Interestingly, the mean difference of attribute strengths is below zero. It implies that SDs tend to omit some concepts
such as \texttt{group}\footnote{e.g., A group of people is standing around a large clock.} or \texttt{plate}\footnote{e.g., A table is set with two plates of food and a candle.}. In particular, SDv2.1 
struggles to generate scenes with multiple people. It aligns with common claims\footnote{https://www.assemblyai.com/blog/stable-diffusion-1-vs-2-what-you-need-to-know/} about SDv2.1 even though it achieves low FID. 
We also conduct similar experiments on the LAION-2B dataset, and SDv1.5 shows continuously better SaD and PaD. We provide more details in Appendix~\ref{subsec:coco_detail}.

\begin{table}[]
\centering
\caption{\textbf{SaD and PaD of different versions of Stable Diffusion.} Stable Diffusion v1.5 is almost twice better than v2.1. We generate 30,000 images using the captions from COCO. We use $\textit{N}_{\mathcal{A}}=30$.}
\resizebox{\linewidth}{!}{
\label{tab:stable}
\begin{tabular}{c|cc|ccc}
\multirow{2}{*}{$\textit{N}_{\mathcal{A}}=30$} & \multirow{2}{*}{SaD ($10^{-7}$)\textdownarrow} & \multirow{2}{*}{PaD ($10^{-7}$)\textdownarrow} & \multicolumn{3}{c}{\shortstack{SaD worst-rank attribute \\ (mean difference)}} \\
                         &                      &                      & 1st              & 2nd              & 3rd               \\ \hline
SDv1.5                   & \textbf{24.37}                & \textbf{60.71}                & \shortstack{plate\\ (-1.9)}     & \shortstack{group\\ (-1.6)}     & \shortstack{building \\(-1.6)}      \\
SDv2.1                   & 48.23               & 106.86               & \shortstack{group \\(-3.7)}     & \shortstack{plate \\(-2.5)}     & \shortstack{person \\(-2.7)}    
\end{tabular}}
\end{table}

\subsection{Impact of sample size and attribute count on proposed metric}
\label{sec:5.5}
\begin{figure}[t]
    \centering
    \includegraphics[width=1\linewidth]{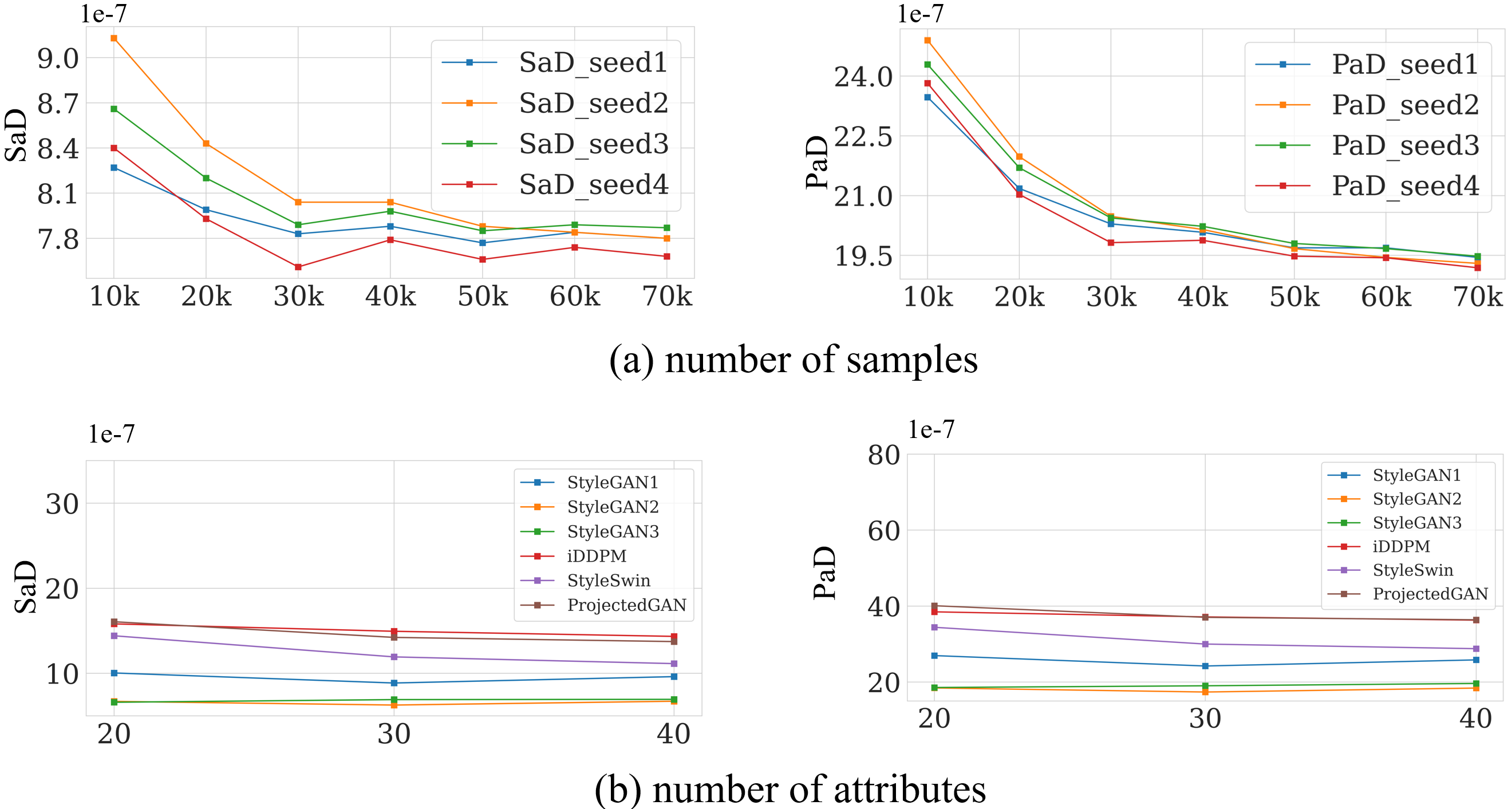}
    \caption{\textbf{SaD and PaD over a different number of samples and attributes.} (a) SaD and PaD are stable with more than 50,000 images. (b) The ranking of models mostly remains consistent regardless of the number of attributes.}
    \vspace{-1em}
    \label{fig:numbers}
\end{figure}

In \fref{fig:numbers}, we conduct ablation experiments to study the impact of the number of samples and attributes. Using 
four random seeds, we generate images with StyleGAN3 from FFHQ. 
We posit that 
SaD and PaD begin to standardize with 30,000 images and become more stable with over 50,000 images.
\fref{fig:numbers}b provides SaD and PaD of various models over different numbers of attributes where the attributes from BLIP are sorted by their number of occurrences in the dataset. The ranking of the models largely stays stable irrespective of the number of attributes. 
We suggest that  20 attributes are sufficient for typical evaluation, but leveraging a broader range offers richer insights.

\subsection{Alignment with human judgment}

Prior metrics \citep{heusel2017gans, sajjadi2018assessing, naeem2020reliable}  faced challenges in evaluating the correspondence with human judgment between large image sets, given the impracticality of storing visual features of thousands of images in human memory. In response, we conduct an alternative approach, assessing the alignment between our metric and human judgment, particularly when there are shifts in attribute distribution(s) within one set.  

\begin{figure}[t]
    \centering
    \includegraphics[width=0.6\linewidth]{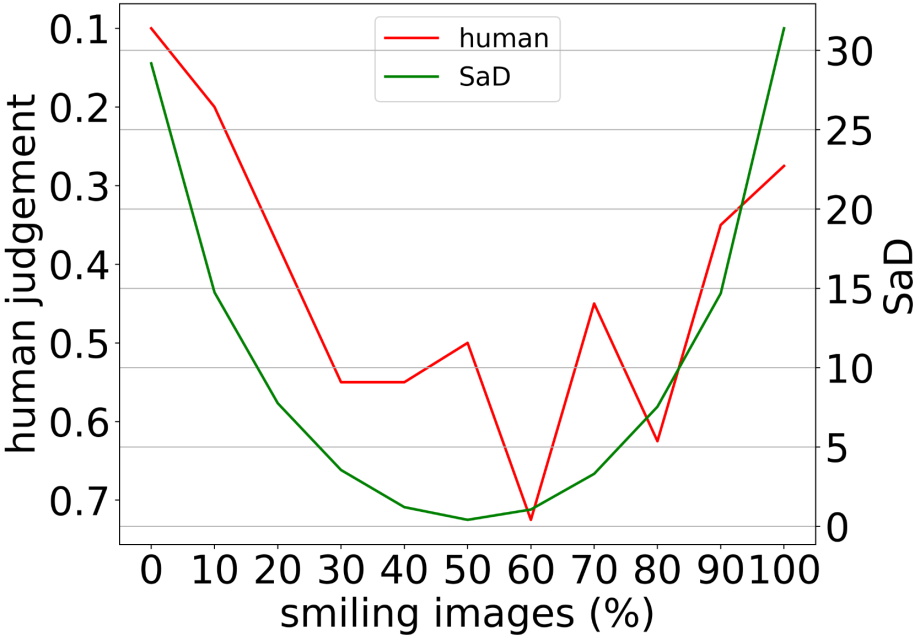}
    \caption{\textbf{Correlation between human judgements and SaD.}}
    \label{fig:sad_human}
\end{figure}

We show SaD and PaD are consistent with human judgment on the CelebA dataset. 40 participants participated in these surveys.
\paragraph{SaD}
\fref{fig:sad_human} shows a correlation between SaD and human judgments. We asked the participants to mark if two sets have different distribution of smile. One set is fixed as a training set with 50\% of them \texttt{smile}. Another set varies from 0\% of them \texttt{smile} to 100\% of them \texttt{smile}. We used smiling and non-smiling images from CelebA ground truth labels. Meanwhile, we measure SaD between the two sets and for comparison. 
Notably, both SaD and human judgement rapidly increase with increasing and decreasing smile in >80\% and <30\% range, respectively. Likewise, both SaD and human judgement have gentle change with same sign of slope in 30\% < smile < 80\% range.
\paragraph{PaD}
\tref{tab:pad_human}  shows a correlation between PaD and human judgments. Based on the given ground-truth set A, participants ranked three sets; 1) a set with strong positive correlation (r=1) 2) a set with zero-correlation (r=0) and 3) a set with strong negative correlation (r=-1).

We opt to use the correlations between \texttt{man} and  \texttt{smile} and we gave five triplets to the participants to rank within the triplets. Most (about 94\%) of the participants identified the rank of correlation between \texttt{man} and \texttt{smile} correctly and it aligns with PaD.


\section{Conclusion and Discussion}

We have introduced novel metrics that evaluate the distribution of attribute strengths. Single-attribute Divergence reveals which attributes are correctly or incorrectly modeled. Paired-attribute Divergence considers the joint occurrence of attributes in individual images. The explicit interpretability of these metrics allows us to know which generative model suits the user's necessity. Furthermore, Heterogeneous CLIPScore more accurately captures the attribute strengths than CLIPScore.

Our metrics have the advantage of revealing the distribution of attributes from a \textit{set} of generated images where human judgment faces difficulty in observing attributes in excessively many images. Furthermore, our research establishes a solid foundation for the development of explainable evaluation metrics for generative models and contributes to the advancement of the field. 

\begin{table}[!t]
\centering
\caption{\textbf{Correlation between human judgements and PaD.}}
\resizebox{1\columnwidth}{!}{%
\begin{tabular}{c|c|c|ccc}
set A & set B      & PaD    & Human 1st (\%) & Human 2nd (\%) & Human 3rd (\%) \\ \hline
\multirow{3}{*}{r=1}   & r=1  & \textbf{4.57}   & \textbf{94.36}         & 3.59          & 2.05          \\
      & r=0  & 38.43  & 2.56          & \textbf{93.85}         & 3.59          \\
      & r=-1 & 117.58 & 3.08          & 2.56          & \textbf{94.36}        
\end{tabular}
}
\label{tab:pad_human}
\end{table}

\paragraph{Discussion} 1) Estimating probability density functions (PDFs) with Kernel Density Estimation (KDE) requires a sufficient ($>$50K) number of samples. With a sufficient number of samples, KDE can effectively approximate the PDF of a dataset, capturing the underlying distribution of the data. This is particularly important in complex data sets where the distribution of attributes might be intricate or not immediately obvious. 2) Our metrics can be influenced by quality of vision-language model (VLM). I.e., a biased or limited extraction of attributes may bring misrepresentation of our metrics. For instance, a VLM that disproportionately emphasizes certain attributes or overlooks others can skew the analysis, leading to an inaccurate assessment of the data. 
3) Exploring strengths of other aspects such as texture \citep{caron2021emerging, oquab2023dinov2, kirillov2023segment} or other modalities \citep{girdhar2023imagebind} may provide valuable insights and enhance the robustness.

Furthermore, we wish to highlight the flexibility of our approach. While we have conducted evaluations using attributes defined by BLIP, our method allows for the customization of attributes to suit the specific needs of the task and the user's objectives. For instance, in pursuing fairness with a focus on equitable generation of features such as race and gender, these can be directly employed as attributes. Alternatively, for specific tasks like image translation, desired attributes can be selectively chosen to tailor the evaluation process. However, it's crucial to exercise caution in this selection process of attributes to avoid introducing bias. 

We believe our new evaluation metrics, with interpretability and the ability to encapsulate user intention, will have a healthy impact on the research community.





\section*{Impact Statement}
This paper presents work whose goal is to advance the field of Machine Learning. There are many potential societal consequences of our work, such as if the training set contains biases, our metrics favor biased generation, which may introduce negative societal impact.

\section*{Acknowledgements}
This work was supported by the National Research Foundation of Korea (NRF) funded by the Korean government (MSIT) (RS-2023-00223062).

\nocite{langley00}

\bibliography{reference_papers}
\bibliographystyle{icml2024}

\newpage
\appendix
\onecolumn


\begin{appendix}
\renewcommand{\thetable}{S\arabic{table}}
\renewcommand{\thefigure}{S\arabic{figure}}

\section{Implementation Details}

\subsection{Additional experiment setup}

\paragraph{Details of generated images}
We generate samples using official checkpoints provided by StyleGANs \citep{karras2019style, karras2020analyzing, karras2020training, karras2021alias}, ProjectedGAN \citep{sauer2021projected}, Styleswin \citep{zhang2022styleswin}, iDDPMs \citep{nichol2021improved, choi2022perception}, and LDM \citep{rombach2022high}. We use 50K of training images and generated images for both FFHQ \citep{karras2019style} and LSUN Cat \citep{yu2015lsun} experiment.

\paragraph{Miscellaneous}
We use \texttt{scipy.stats.gaussian\_kde}(dataset, `\textit{scott}', \text{None}) to estimate the distribution of \dcs{} for given attributes. We observe  HCS values mainly feature unimodal and bimodal distributions, ensuring that the sample sizes are large enough for effective KDE application. We've chosen Scott's Rule for bandwidth selection, as recommended by the default settings in  \texttt{scipy.stats.gaussian\_kde}. This recommendation is due to its balanced approach to managing bias and variance in our data estimation, adjusting the bandwidth based on data size and dimensionality. In KDE, the bandwidth directly influences the standard deviation of the Gaussian kernels; a larger bandwidth leads to a smoother density estimate, while a smaller bandwidth results in a more detailed density estimate. This directly affects the std of the kernels used in our analysis. This rule scales the bandwidth with $n^{-1/(d+4)}$ , where $n$ is the number of data points, and $d$ is the number of dimensions.

We use \texttt{spacy.load("en\_core\_web\_sm")} to extract attributes from BLIP\citep{li2022blip} captions.
We resize all images to 224x224. We used \texttt{"ViT-B/32"} \citep{dosovitskiy2020image} as a CLIP encoder. We used a single NVIDIA RTX 3090 GPU (24GB) for the experiments.

\subsection{Details of CelebA accuracy experiment}
\label{sec:accuracy}

\tref{tab:DSC_from_img_mean} displays binary classification results for all attributes in CelebA using both CS and HCS, comparing them to the ground truth attribute labels. By setting the threshold based on the number of positive labels for each CelebA attribute, we found that the accuracy and F1 score of HCS are superior to CS, regardless of whether we use micro or macro averaging. Additionally, we conducted experiments by setting the origin of HCS as the overall mean of both image and text means, validating that using separate text and image means is essential.

\begin{table}[h]
\caption{\textbf{Attributes used for CelebA accuracy experiment.}}
\label{tab:celebA_refine_attributes}
\resizebox{0.9\textwidth}{!}{
\begin{tabular}{c|c}
Attribute type     & Attribute                                                                                                                                                                                                                                                                                       \\ \hline
Refined attributes & \begin{tabular}[c]{@{}c@{}}Arched\_Eyebrows, Bags\_Under\_Eyes, Bald, Bangs, Big\_Nose, \\ Black\_Hair, Blond\_Hair,Brown\_Hair, Chubby, Double\_Chin, Eyeglasses, \\ Goatee, Gray\_Hair, Heavy\_Makeup, Male, Mouth\_Slightly\_Open, Mustache, \\ No\_Beard, Sideburns, Smiling, Straight\_Hair, Wavy\_Hair, Wearing\_Earrings, \\ Wearing\_Hat,Wearing\_Lipstick, Wearing\_Necklace, Wearing\_Necktie, Young\end{tabular}                                                                                                                                                                                          \\ \hline
All attributes     & \begin{tabular}[c]{@{}c@{}}5\_o\_Clock\_Shadow, Arched\_Eyebrows, Attractive, Bags\_Under\_Eyes, \\ Bald, Bangs, Big\_Lips,Big\_Nose, Black\_Hair, Blond\_Hair,  Blurry, \\ Brown\_Hair, Chubby, Double\_Chin, Eyeglasses, Goatee, Gray\_Hair, \\ Heavy\_Makeup, High\_Cheekbones, Male, Mouth\_Slightly\_Open, Mustache, \\ Narrow\_Eyes, No\_Beard, Oval\_Face, Pale\_Skin, Pointy\_Nose, Receding\_Hairline, \\ Rosy\_Cheeks, Sideburns, Smiling, Straight\_Hair, Wavy\_Hair, Wearing\_Earrings, \\ Wearing\_Hat,Wearing\_Lipstick, Wearing\_Necklace, Wearing\_Necktie, Young\end{tabular}
\end{tabular}
}

\end{table}

\begin{table}[!t]
\centering
\caption{\textbf{Accuracy from CelebA ground truth labels.} 
\dcs{} with origin at the entire center of images and texts is seriously inferior to the one with origin at the separate center of images ($C_\mathcal{X}$) and texts ($C_\mathcal{A}$). It validates the definition of~$V_a$.}

\begin{tabular}{c|c|ccc}
                                    &               & accuracy       & f1 score(macro) & f1 score(micro) \\ \hline
\multirow{2}{*}{All attributes}     & HCS & \textbf{0.794} & \textbf{0.442}  & \textbf{0.545}  \\
                                    & CS            & 0.781          & 0.392           & 0.515           \\ \hline
\multirow{2}{*}{Refined attributes} & HCS & \textbf{0.817} & \textbf{0.519}  & \textbf{0.616}  \\
                                    & CS            & 0.798          & 0.450           & 0.575          
\end{tabular}
\label{tab:DSC_from_img_mean}
\end{table}


\begin{table}[!t]
\centering
\caption{\textbf{Top 40 appeared attribute in COCO validation captions.} The first and third rows represent the attributes in COCO validation captions, while the second and fourth rows represent the corresponding number of appearances of these attributes in the captions.}
\resizebox{0.75\textwidth}{!}{
\begin{tabular}{c|c|c|c|c|c|c|c|c|c}
man   & woman   & he         & people  & person & table & group    & street & water     & plate  \\
20262 & 9352    & 8212       & 8164    & 7196   & 6584  & 6401     & 4382   & 3741      & 3717   \\ \hline
cat     & field & couple & dog   & side & food & beach & bed   & bathroom & road \\
3476    & 3385  & 3301   & 3071  & 2981 & 2973 & 2731  & 2687  & 2477     & 2377 \\ \hline
grass & kitchen & skateboard & picture & road   & train & building & snow   & surfboard & toilet\\
2346  & 2286    & 2259       & 2209    & 2165   & 2140  & 2108     & 2097   & 1968  &1879    \\ \hline
giraffe & room  & men    & bunch & ball & air  & bench & clock & boy      & sign \\
1874    & 1827  & 1819   & 1809  & 1807 & 1710 & 1630  & 1607  & 1573     & 1569\\
\end{tabular}}
\label{tab:coco_caption}
\end{table}

\section{Additional Ablation Study}

\subsection{Necessity of separating image mean and text mean}
In the main paper, we defined \dcs{} as computing angles between vectors $V_x$ and $V_a$. $V_x$ is a vector from the center of images to an image in CLIP space. $V_a$ is a vector from the center of captions to an attribute in CLIP space.
\tref{tab:DSC_from_img_mean} quantitatively validates the effectiveness of setting the origin of $V_a$ as the center of captions ($C_\mathcal{A}$) compared to the center of images ($C_\mathcal{X}$).

\subsection{Replacing \dcs{} with \cs{}}
We also include additional comparisons of \oneabbr{} and \twoabbr{} across different image injection settings with \cs{} rather than \dcs{} in \fref{fig:injection}. Compared to the validation result with \dcs{}, both results reflect a corresponding tendency: the more correlated image injected, the worse performance in the proposed metric. However, considering the quantitative effectiveness we demonstrated for \dcs{} in \tref{tab:DSC_from_img_mean}, we highly recommend using \dcs{} with proposed metrics: \oneabbr{} and \twoabbr{}.

\begin{figure}[t]
    \centering
    \includegraphics[width=1\linewidth]{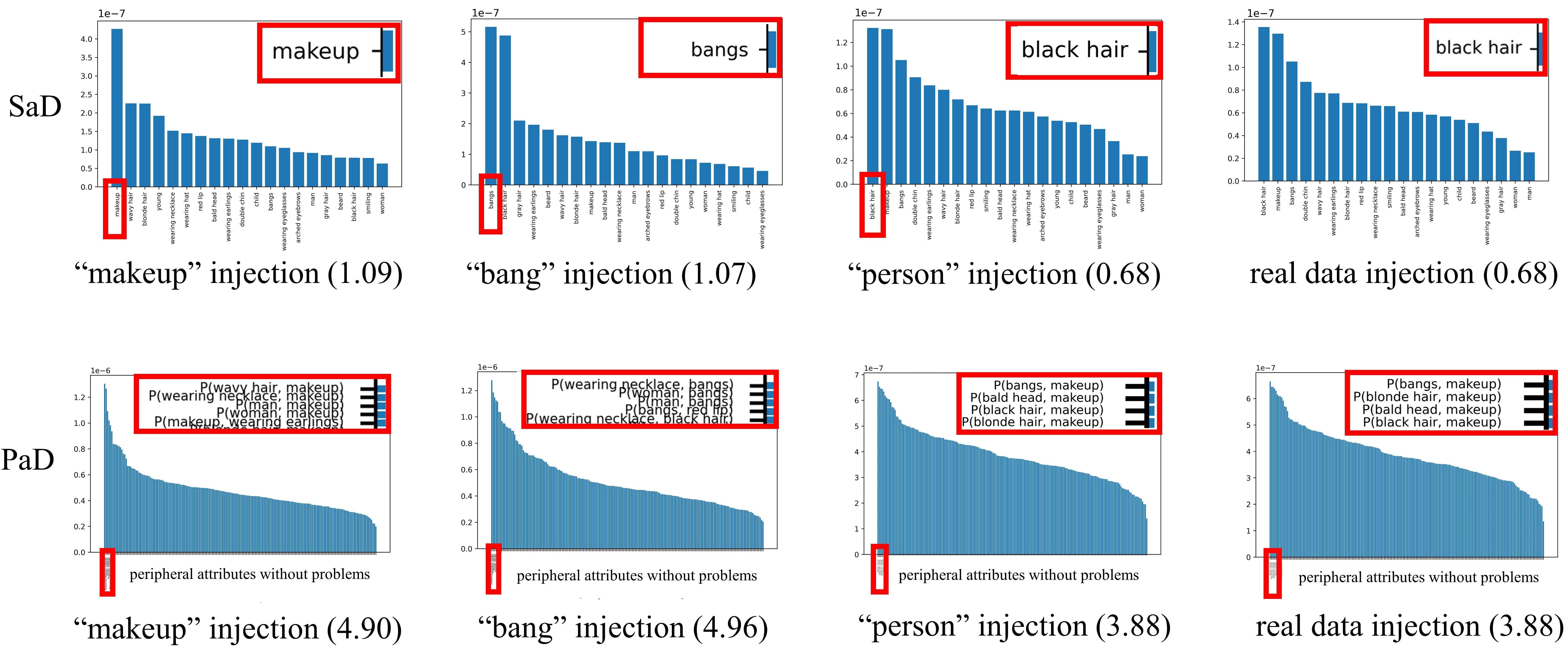}
    \caption{\textbf{Correlated images injection experiment.}} \vspace{-0.5em}
    \label{fig:injection_appendix}
\end{figure}

\begin{figure}[t]
    \centering
    \includegraphics[width=1\linewidth]{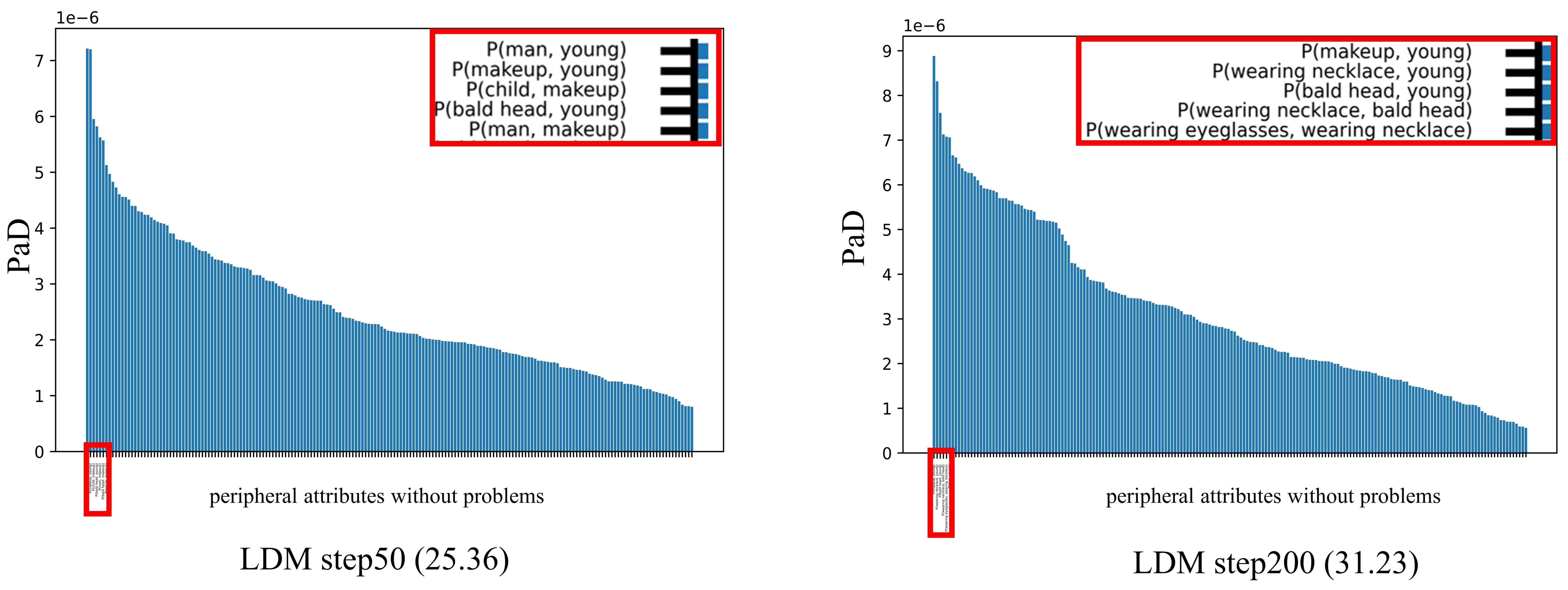}
    \caption{\textbf{PaD for LDM with different sampling timesteps.}} \vspace{-0.5em}
    \label{fig:ldm_pad}
\end{figure}


\begin{table}[!t]
\centering
\caption{\textbf{SaD and PaD for text-to-image models.} Stable diffusion v1.5 outperforms Stable Diffusion v2.1 in SaD and PaD regardless of the number of attributes despite being known as inferior in FID.}
\begin{tabular}{l|ccc|ccc}
       & \multicolumn{3}{c|}{SaD} & \multicolumn{3}{c}{PaD}  \\ \hline
       & $\textit{N}_{\mathcal{A}}=20$     & $\textit{N}_{\mathcal{A}}=30$     & $\textit{N}_{\mathcal{A}}=40$     & $\textit{N}_{\mathcal{A}}=20$     & $\textit{N}_{\mathcal{A}}=30$     & $\textit{N}_{\mathcal{A}}=40$     \\
SDv1.5 & \textbf{37.91}  & \textbf{24.37}  & \textbf{25.44}  & \textbf{87.53}  & \textbf{60.71}  & \textbf{62.47}  \\
SDv2.1 & 69.49  & 48.23  & 47.53  & 146.12 & 106.86 & 105.03
\end{tabular}
\label{tab:pad_stable}
\end{table}

\subsection{Attribute combination involving more than two attributes}
 Proposed metric is capable of examining relationships involving any number of attributes (N-way relationships), as it fundamentally relies on measuring the joint probability among involved attributes. The methodology proposed in our paper remains applicable for considering the joint probability of three or more attributes. This is particularly true for exceeding 30k images, where we did not observe statistical instability.

While it's technically feasible to extend the analysis to encompass interactions among three or more attributes (N-way relationships), we focus on the two-attribute relationships, which provide the finest level of granularity. Furthermore, our experiments have shown that the model rankings remains consistent regardless of the complexity of attribute relationships considered.

\tref{tab:3attr} supports the above statement: the left two columns report the worst three triplets of attributes and the rightmost column reports the worst pairs with their ranks connected to each triplet. The pairs being included in the triplets in the same row indicates that the worst triplets can be identified by the worst pairs. 

\begin{table}[!t]
\centering
\caption{\textbf{Results with attribute triplets}}
\begin{tabular}{c|cc}
triplet ranks & triplet attributes           & similar PaD ranks (PaD attributes)    \\ \hline
1             & man \& woman \& wearing necklace & R1(man\& woman)                         \\
2             & child \& red lip \& makeup       & R2(red lip\& makeup), R3(child \& makeup) \\
3             & red lip \& makeup \& young       & R2(red lip\& makeup), R4(makeup \& young)
\end{tabular}
\label{tab:3attr}
\end{table}

\subsection{Can SaD and PaD also capture skips of attribute?}
\label{subsec:skip}
We validate that SaD and PaD accurately capture the skipness of certain attributes in \tref{tab:eyeglass}. Using CelebA annotation labels, we construct sets A and B with 50k images, each naturally containing 3,325 and 3,260 images with eyeglasses, respectively. As we intentionally replace images with eyeglasses in set B with images without eyeglasses,  SaD and PaD deteriorated linearly with an increasing number of replaced images, with the \texttt{eyeglasses} attribute making a more significant contribution to SaD and PaD. It demonstrates proposed metric effectively catches the skipness of some attributes, and accurately captures the distribution change of the attribute HCS probability density function.

\begin{table}[]
\centering
\caption{\textbf{Validation result of skips experiment.}} 
\renewcommand{\arraystretch}{1.5}
\begin{tabular}{l|ccc}
                                                              & SaD  & PaD   & most influencing attribute for SaD \\ \hline
$\text{eyeglasses 3325} \over \text{total 50000}$ v.s. $\text{eyeglasses 3260} \over \text{total 50000}$  & 0.63 & 3.42  & beard                              \\
$\text{eyeglasses 3325} \over \text{total 50000}$  v.s. $\text{eyeglasses 2000} \over \text{total 50000}$  & 0.89 & 4.05  & \textbf{eyeglasses}                \\
$\text{eyeglasses 3325} \over \text{total 50000}$  v.s. $\text{eyeglasses 1000} \over \text{total 50000}$  & 1.54 & 5.66  & \textbf{eyeglasses}                \\
$\text{eyeglasses 3325} \over \text{total 50000}$  v.s. $\text{eyeglasses 3325} \over \text{total 50000}$     & 3.25 & 11.59 & \textbf{eyeglasses}               
\end{tabular}
\label{tab:eyeglass}
\end{table}

\subsection{More details: text-to-image model evaluation}
\label{subsec:coco_detail}
\paragraph{The number of attributes}
We compare Stable Diffusion v1.5 and Stable Diffusion v2.1 on the COCO dataset using top-N appeared attributes in COCO validation captions (\tref{tab:coco_caption}). Regardless of number of attributes, SDv1.5 outperforms SDv2.1 in SaD and PaD (\fref{fig:coco_sad}, \tref{tab:pad_stable}). 

\paragraph{Different dataset} We compare Stable Diffusion v1.5 and Stable Diffusion v2.1 with a 30k subset of the LAION-2B dataset as shown in \tref{tab:laion}. The outcomes align closely with the values reported in the paper.

\begin{table}[!t]
\centering
\caption{\textbf{SaD and PaD for text-to-image models: LAION-2B.} Stable diffusion v1.5 outperforms Stable Diffusion v2.1 in SaD and PaD regardless of dataset type despite being known as inferior in FID.}
\begin{tabular}{c|cc}
       & SaD            & PaD            \\ \hline
SDv1.5 & \textbf{14.26} & \textbf{40.27} \\
SDv2.1 & 32.30          & 62.54         
\end{tabular}
\label{tab:laion}
\end{table}

\begin{figure}[t]
    \centering
    \includegraphics[width=0.8\linewidth]{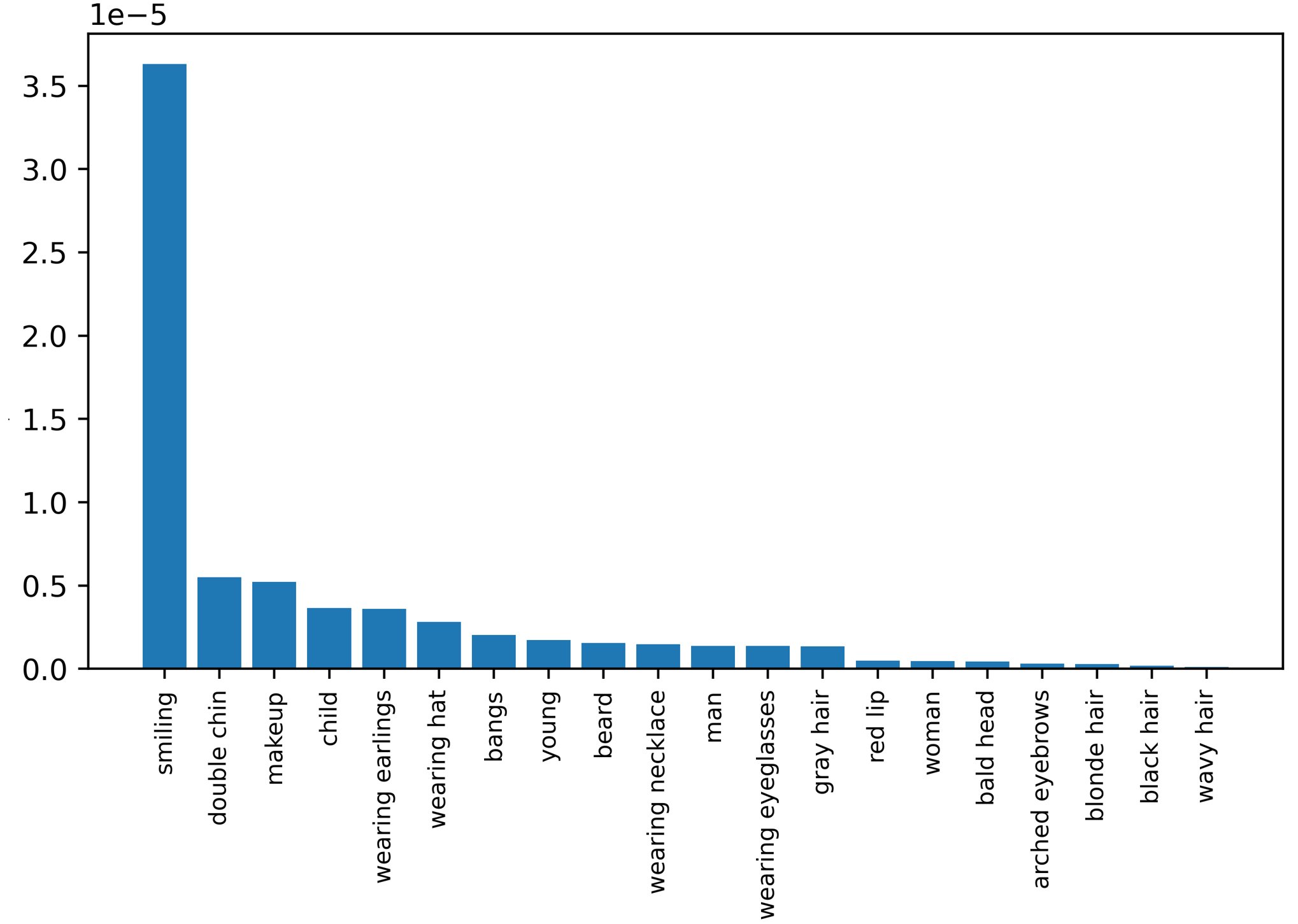}
    \caption{\textbf{SaD result for \sref{sec:5.2} (Discernment of PaD).}}
    \vspace{-0.5em}
    \label{fig:5_2_SaD}
\end{figure}

\begin{figure}[t]
    \centering
    \includegraphics[width=0.8\linewidth]{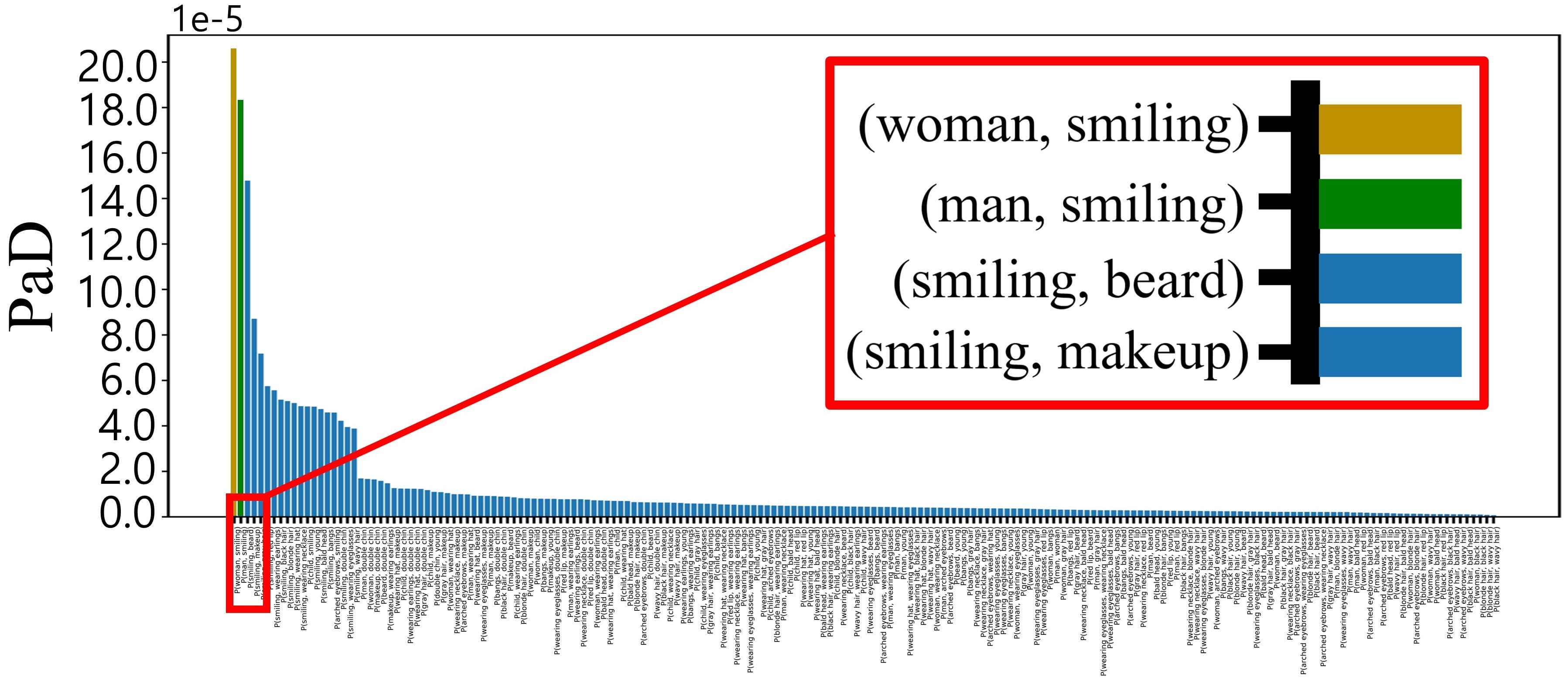}
    \caption{\textbf{PaD for \sref{sec:5.2}: Discernment of PaD.}}
    \vspace{-0.5em}
    \label{fig:5_2_PaD}
\end{figure}

\begin{figure}[t]
    \centering
    \includegraphics[width=0.9\linewidth]{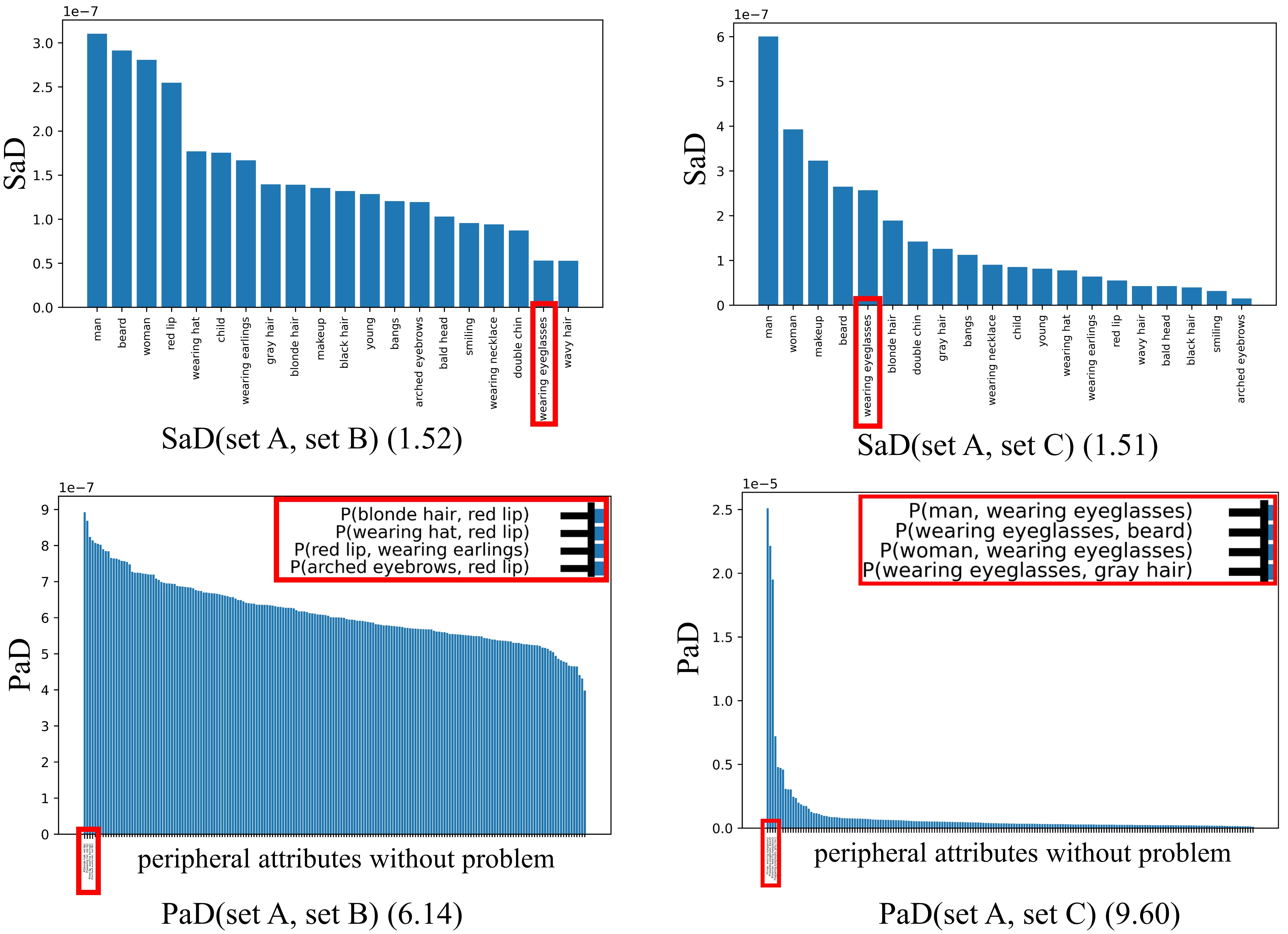}
    \caption{\textbf{Additional experiment for \sref{sec:5.2}: Necessity of PaD over SaD.} In set A and set B only \texttt{men} wear \texttt{eyeglasses}, while only \texttt{women} wear \texttt{eyeglasses} in set C. PaD successfully captures pairwise relation errors between set A and set C, whereas SaD cannot.}
    \vspace{-0.5em}
    \label{fig:eyeglass_100}
\end{figure}

\begin{table}[!t]
\centering
\caption{\textbf{Correlation between human judgements and SaD.}}
\begin{tabular}{c|c|c|ccc}
set A        & set B        & SaD   & Human 1st (\%) & Human 2nd (\%) & Human 3rd (\%) \\ \hline
\multirow{3}{*}{strong smile} & strong smile & \textbf{0.89}  & \textbf{99.48}     & 0             & 0.51          \\
             & medium smile & 19.39 & 0             & \textbf{99.48}     & 0.51          \\
             & no smile     & 92.46 & 0.51          & 0.51          & \textbf{98.97}    
\end{tabular}
\label{tab:sad_human}

\end{table}

\begin{table}[!t]
\centering
\caption{\textbf{Alignment of result from different attribute selection methologies.} Regardless of attribute selection methologies (BLIP, CelebA GT labels), tendency of SaD and PaD remains same.}
\resizebox{\textwidth}{!}{
\begin{tabular}{c|c|cccccccc}
                                 &     & StyleGAN1 & StyleGAN2      & StyleGAN3     & iDDPM & LDM (50) & LDM (200) & StyleSwin & ProjectedGAN \\ \hline
\multirow{2}{*}{BLIP}            & SaD ($10^{-7}$)\textdownarrow  & 10.04     & 6.70           & \textbf{6.60} & 15.81 & 12.97    & 10.36     & 14.41     & 16.06        \\
                                 & PaD ($10^{-7}$)\textdownarrow  & 26.96     & \textbf{18.44} & 18.56         & 38.48 & 31.87    & 26.03     & 34.41     & 40.09        \\ \hline
\multirow{2}{*}{CelebA GT label} & SaD ($10^{-7}$)\textdownarrow  & 11.35     & \textbf{7.52}  & 7.79          & 14.78 & 10.42    & 14.04     & 10.76     & 17.61        \\
                                 & PaD ($10^{-7}$)\textdownarrow  & 27.25     & \textbf{19.22} & 19.73         & 34.04 & 25.36    & 30.71     & 26.56     & 41.53       
\end{tabular}}
\label{tab:allignment}

\end{table}

\renewcommand{\thetable}{S\arabic{table}}
\renewcommand{\thefigure}{S\arabic{figure}}

\section{Is BLIP enough to represent attributes of images?}

\tref{tab:allignment} shows the alignment between the results extracted using BLIP and the CelebA GT labels used as attributes. Through this, we argue that the attributes extracted by BLIP can adequately represent the significant attributes of an image dataset. Additionally, we can select specific domain attributes using LLMs. In the case of LSUN, we used LLMs to extract attributes related to shape and texture.

\paragraph{Details of GPT queries}
\label{sec:gpt}
\tref{tab:gpt} provides the questions we used for preparing GPT attributes.
We accumulated GPT attributes by iteratively asking GPT to answer `Give me 50 words of useful, and specific adjective visual attributes for \{\textit{question}\}'. Then, we selected the top N attributes based on their frequency of occurrence, ensuring that the most frequently mentioned attributes were prioritized. We suppose that the extracted attributes might be biased due to the inherent randomness in GPT's answering process. This potential problem is out of our scope. We anticipate future research will address it to extract attributes in a more fair and unbiased manner with large language models. For a smooth flow of contents, the table is placed at the end of this material.

\begin{table}[h]
\caption{\textbf{Scripts used for extracting attributes from GPT.} We stack GPT attributes by iteratively asking GPT to answer `Give me 50 words of useful, and specific adjective visual attributes for \{\textit{question}\}'.}
\centering
\begin{tabular}{c|l}
Dataset  & \multicolumn{1}{c}{{\textit{question}}}                                                                                                                                                                                        \\ \hline
FFHQ     & \begin{tabular}[c]{@{}l@{}}`distinguishing faces in a photo'\\ `distinguishing human faces in a photo'\\ `distinguishing different identities of people in photos of faces'\\ `differentiating between people's faces by their distinctive features'\\ `people to change there styles in hairs, accessories around their faces'\\ `recognizing changes in hair and accessory styles in photographs of people's faces'\\ `identifying distinct faces within an image\\ `recognizing facial characteristics to distinguish people in photos'\\ `discerning variations in facial features to identify people in images'\\ `spotting differences in facial appearance for identifying individuals'\end{tabular}                                                                                                                                                      \\ \hline
LSUN Cat & \begin{tabular}[c]{@{}l@{}}` recognizing individuals from facial features in photographs'\\ `identifying distinct faces within an image\\ `recognizing variations in feline appearance to identify individual cats'\\ `discerning differences in fur patterns and colors to distinguish cats in photos'\\ `detecting subtle facial expressions to distinguish emotions in cat photos'\\ `differentiating between cats based on body type and size in photos'\\ `identifying distinctive facial features to distinguish between cats in images'\\ `recognizing changes in coat texture and length in photos of cats'\\ `discerning variations in eye color and shape to identify individual cats in images'\\ `spotting unique markings to distinguish between cats in photos'\end{tabular}                                                                        \\ \hline
\end{tabular}

\label{tab:gpt}
\end{table}

\paragraph{Details of extracted attribute}
\tref{tab:attr_list} describes selected attributes by each extractor. We used "A photo of \{attribute\}" as prompt engineering for all attributes.

\begin{table}[h]
\caption{\textbf{Examples of extracted attributes by each attribute extractors.}}
\label{tab:attr_list}
\centering

\begin{tabular}{c|c|c}
Extractor             & N  & Attribute                                                                                                           \\ \hline

\multirow{14}{*}{BLIP} & 20 & \begin{tabular}[c]{@{}c@{}}woman, man, person, glasses, suit, \\ little girl, tie, picture, sunglasses, young boy, \\ cell phone, microphone, necklace, hat, young girl, \\ blonde hair, long hair, blue shirt, beard, white shirt\end{tabular} \\ \cline{2-3} 
& 30 & \begin{tabular}[c]{@{}c@{}}
woman, man, person, glasses, suit, \\ little girl, tie, picture, sunglasses, young boy, \\ cell phone, microphone, necklace, hat, young girl, \\ blonde hair, long hair, blue shirt, beard, white shirt, \\ her head, her face, couple, baby, her hair, \\ scarf, black shirt, smile, young man, little boy, child \end{tabular} \\ \cline{2-3}
& 40 & \begin{tabular}[c]{@{}c@{}}
woman, man, person, glasses, suit, \\ little girl, tie, picture, sunglasses, young boy, \\ cell phone, microphone, necklace, hat, young girl, \\ blonde hair, long hair, blue shirt, beard, white shirt, \\ her head, her face, couple, baby, her hair, \\ scarf, black shirt, smile, young man, little boy, child, \\ red hair, flower, her hand, his mouth, blue eyes, women\end{tabular} \\ \hline

\multirow{15}{*}{GPT}  & 20 & \begin{tabular}[c]{@{}c@{}}clean-shaven, beard, mustache, wide-eyed, thin lips,\\ bald, glasses-wearing, freckled, almond-shaped eyes,\\  scarred, wrinkled, soul patch, high forehead, hooded eyes, \\ piercings, prominent cheekbones, full lips, \\ braided, upturned-nosed, youthful\end{tabular}                                                                                                                                                                                                                                                                                                                                                     \\ \cline{2-3} 
                      & 30 & \begin{tabular}[c]{@{}c@{}}clean-shaven, beard, mustache, wide-eyed, thin lips,\\bald, glasses-wearing, freckled, almond-shaped eyes,\\  scarred, wrinkled, soul patch, high forehead, \\ hooded eyes, piercings, prominent cheekbones, full lips,\\  braided, upturned-nosed, youthful, approachable, \\ arched eyebrows,thin-lipped, thin-eyebrowed, birthmark, \\ bobbed, composed, curly hair, deep-set eyes, thick-eyebrowed\end{tabular}                                                                                              \\ \cline{2-3} 
                      & 40 & \begin{tabular}[c]{@{}c@{}}clean-shaven, beard, mustache, wide-eyed, thin lips,\\ bald, glasses-wearing, freckled, almond-shaped eyes,\\  scarred, wrinkled, soul patch, high forehead,\\  hooded eyes, piercings, prominent cheekbones, full lips,\\  braided, upturned-nosed, youthful, approachable,\\  arched eyebrows,thin-lipped, thin-eyebrowed, birthmark, \\ bobbed, composed, curly hair, deep-set eyes, thick-eyebrowed, \\ earrings, eyebrow thickness, facial hair, goatee, \\ heart-shaped face, long eyelashes, low forehead, \\ monolid eyes, nasolabial folds,  diamond-shaped face\end{tabular} \\ \hline
CelebA GT label                  & 20 & \begin{tabular}[c]{@{}c@{}}makeup, bangs, wearing eyeglasses, wearing earrings,\\  black hair, arched eyebrows, blonde hair, red lip, \\ gray hair, beard, wavy hair, child, bald head, \\ smiling, double chin, wearing hat, young, man,\\  woman, wearing necklace\end{tabular}                                                   
\end{tabular}

\end{table}

\section{More detailed results and analysis}
\label{sec:appen_analysis}

In this section, we provide analysis of various generative models using our metric's explicit interpretability.

\begin{table}[!h]
\caption{\textbf{Top 3 \twoabbr{} pair with USER attributes on FFHQ.}}
\label{tab:pad}
\resizebox{\textwidth}{!}{
\begin{tabular}{l|llllllll}
    & StyleGAN1                                               & StyleGAN2                                                         & StyleGAN3                                                        & iDDPM                                                            & LDM (50)                                                 & LDM (200)                                                          & StyleSwin                                                  & ProjectedGAN                                              \\ \hline
1st & \begin{tabular}[c]{@{}l@{}}man\\ \&woman\end{tabular}    & \begin{tabular}[c]{@{}l@{}}arched\\\ eyebrows\\ \&makeup\end{tabular} & \begin{tabular}[c]{@{}l@{}}red lip\\ \&makeup\end{tabular}        & \begin{tabular}[c]{@{}l@{}}arched\\\ eyebrow\\ \&makeup\end{tabular} & \begin{tabular}[c]{@{}l@{}}man\\ \&young\end{tabular}    & \begin{tabular}[c]{@{}l@{}}makeup\\ \&young\end{tabular}           & \begin{tabular}[c]{@{}l@{}}makeup\\ \&young\end{tabular}    & \begin{tabular}[c]{@{}l@{}}man\\ \&woman\end{tabular}      \\ \hline
2nd & \begin{tabular}[c]{@{}l@{}}child\\ \&makeup\end{tabular} & \begin{tabular}[c]{@{}l@{}}child\\ \&makeup\end{tabular}           & \begin{tabular}[c]{@{}l@{}}arched\\ \ eyebrow\\ \&makeup\end{tabular} & \begin{tabular}[c]{@{}l@{}}woman\\ \&arched\\ \ eyebrow\end{tabular}  & \begin{tabular}[c]{@{}l@{}}makeup\\ \&young\end{tabular} & \begin{tabular}[c]{@{}l@{}}wearing\\ \ necklace\\ \&young\end{tabular} & \begin{tabular}[c]{@{}l@{}}woman\\ \&young\end{tabular}     & \begin{tabular}[c]{@{}l@{}}red lip\\ \&makeup\end{tabular} \\ \hline
3rd & \begin{tabular}[c]{@{}l@{}}makeup\\ \&young\end{tabular} & \begin{tabular}[c]{@{}l@{}}man\\ \&woman\end{tabular}              & \begin{tabular}[c]{@{}l@{}}child\\ \&makeup\end{tabular}          & \begin{tabular}[c]{@{}l@{}}child\\ \&makeup\end{tabular}          & \begin{tabular}[c]{@{}l@{}}child\\ \&makeup\end{tabular} & \begin{tabular}[c]{@{}l@{}}bald head\\ \&young\end{tabular}        & \begin{tabular}[c]{@{}l@{}}wavy hair\\ \&young\end{tabular} & \begin{tabular}[c]{@{}l@{}}child\\ \&makeup\end{tabular}  
\end{tabular}
}
\end{table}

\begin{figure}[t]
    \centering
    \includegraphics[width=1\linewidth]{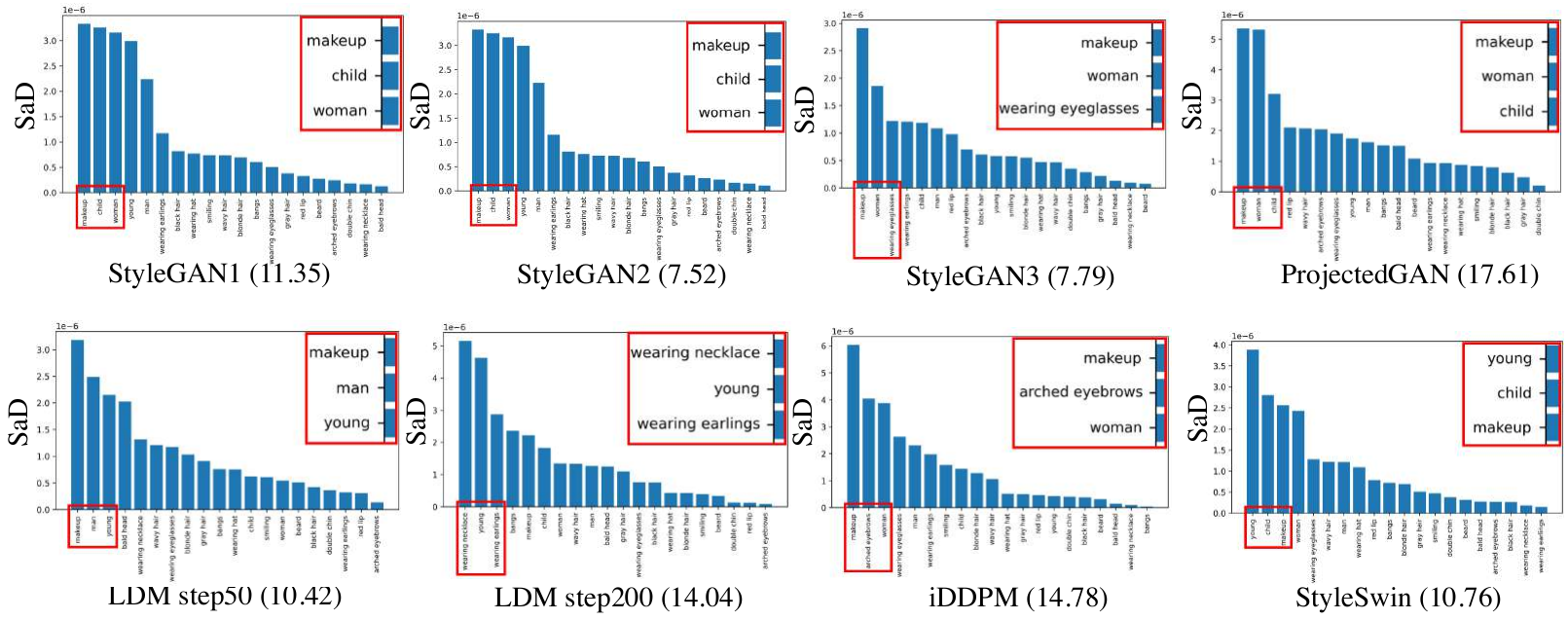}
    \caption{\textbf{SaD with USER attributes on FFHQ.} \vspace{-0.5em}}
    \label{fig:ldm_sad}
\end{figure}

\begin{figure}[t]
    \centering
    \includegraphics[width=1\linewidth]{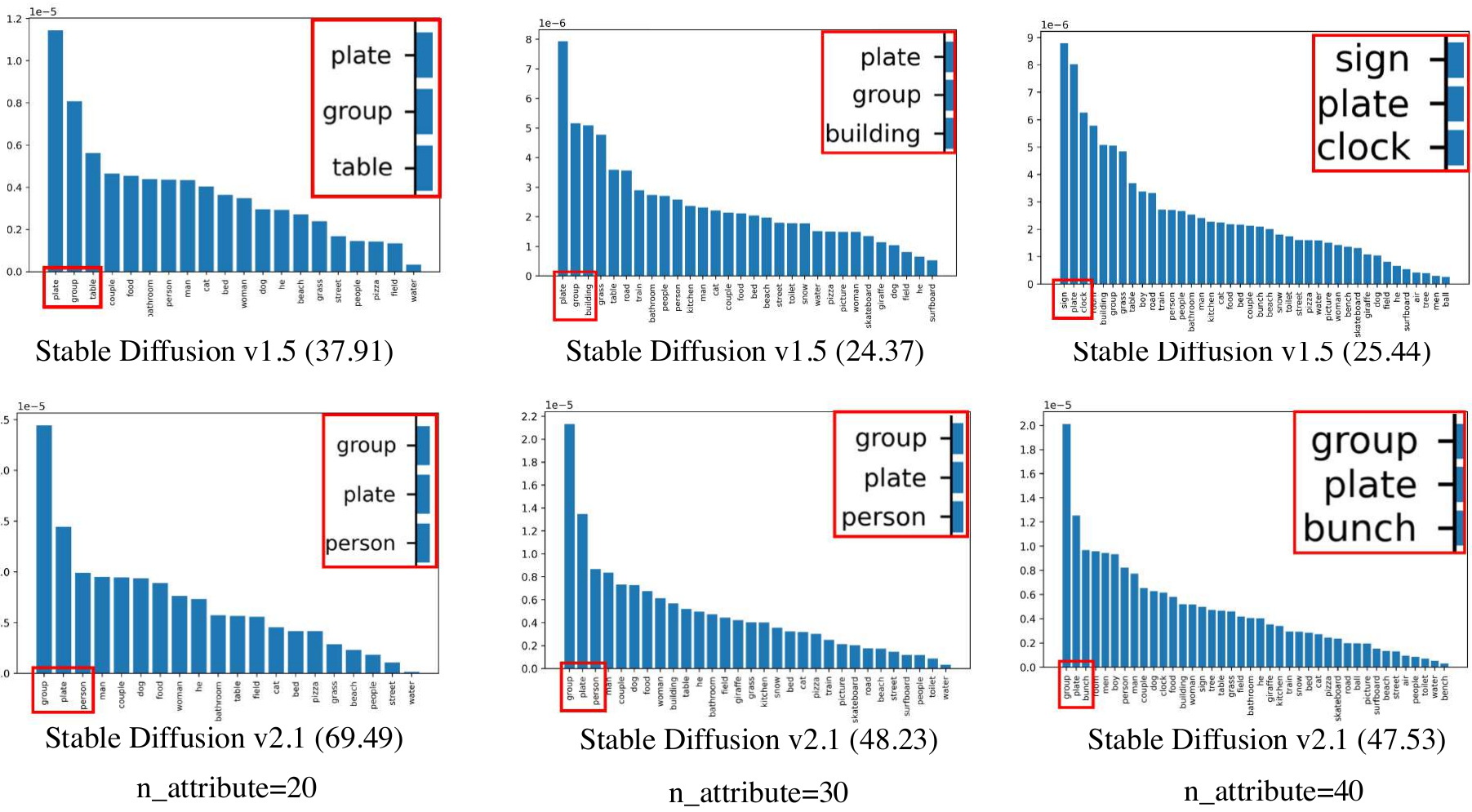}
    \caption{\textbf{SaD for text-to-image models.} Stable Diffusion v1.5 outperforms stable Diffusion v2.1 in SaD. regardless of the number of attributes despite being known as inferior in FID.} \vspace{-0.5em}
    \label{fig:coco_sad}
\end{figure}

\paragraph{SaD}
\fref{fig:ldm_sad} shows the SaD results for StyleGAN 1, 2, 3, iDDPM, and LDM with two different step versions, StyleSwin, and ProjectedGAN. For LDM, DDIM sampling steps of 50 and 200 were used, and all numbers of the images are 50k.

SaD directly measures the differences in attribute distributions, indicating the challenge for models to match the density of the highest-scoring attributes to that of the training dataset. Examining the top-scoring attributes, all three StyleGAN models have similar high scores in terms of scale. However, there are slight differences, particularly in StyleGAN3, where the distribution of larger accessories such as \texttt{eyeglasses} or \texttt{earrings} differs. Exploring the training approach of alias-free modeling and its relationship with such accessories would be an interesting research direction.

In contrast, iDDPM demonstrates notable scores, with attributes \texttt{makeup} and \texttt{woman} showing scores over two times higher than GANs. Particularly, apart from these two attributes, the remaining attributes are similar to GANs, highlighting significant differences in the density of \texttt{woman} and \texttt{makeup}. Investigating how the generation process of diffusion models, which involves computing gradients for each pixel, affects attributes such as \texttt{makeup} and \texttt{woman} would be an intriguing avenue for future research.

For LDM, while FID improves with more timesteps, SaD gets worse. Specifically, the scores for \texttt{earrings}, \texttt{necklace}, and \texttt{young} significantly increase with 200-step results. Analyzing the influence of attributes as the number of steps increases, leading to more frequent gradient updates, would be a highly interesting research direction. Moreover, diffusion models are known to generate different components at each timestep. Understanding how these model characteristics affect attributes remains an open question and presents an intriguing area for exploration.

\paragraph{PaD}

PaD provides a quantitative measure of the appropriateness of relationships between attributes. Thus, if a model generates an excessive or insufficient number of specific attributes, it affects not only SaD but also PaD. Therefore, it is natural to expect that attribute pairs with high PaD scores will often include worst-ranking attributes in SaD. \tref{tab:pad} presents the worst three attributes with the highest PaD scores, and their overall values can be found in \tref{tab:gan_diff_blip}.

PaD reveals interesting findings. Firstly, it is noteworthy that attributes related to \texttt{makeup} consistently receive high scores across all StyleGAN 1, 2, and 3 models. (\tref{tab:pad}) This indicates that GANs generally fail to learn the relationship between \texttt{makeup} and other attributes, making it an intriguing research topic to explore the extent of this mislearning and its underlying reasons.

In the case of iDDPM, the values for \texttt{arched eyebrows} and \texttt{makeup} are overwhelmingly higher compared to other attributes. The reasons behind this will be discussed in the following subsection.

\paragraph{Comparing generative models with specific attribute types} 

\label{sec:appenlsun}
In the main paper, we suppose that the distribution of color-related attributes has a harmful effect on the DMs' performances compared to shape-related attributes on the proposed metric. In this section, we analyze which specific attribute DMs are hard to generate compared to StyleGAN models. 

\begin{figure}[t]
    \centering
    \includegraphics[width=1\linewidth]{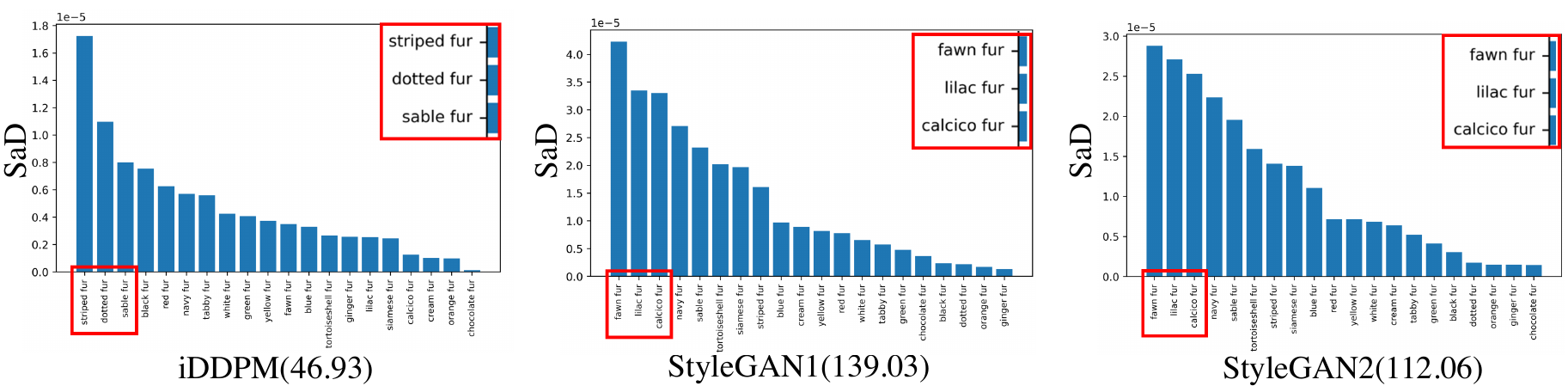}
    \caption{\textbf{SaD for LSUN Cat with color attributes.}}
    \label{fig:lsun_color}
\end{figure}

\paragraph{Color-related attributes}
\fref{fig:lsun_color} illustrates the color-related result of \oneabbr{} that iDDPM fails to preserve attributes with patterns such as \texttt{striped fur} and \texttt{dotted fur}. 
Considering that the color in the diffusion model is largely determined by the initial noise, we suppose that creating texture patterns such as stripes or dot patterns would be challenging. This characteristic is also observed in \twoabbr{}. Unlike GANs, we can observe that relationships between solid colors without patterns or textures are not among the worst 3 attributes. (\tref{tab:lsun_pad}) 

\begin{figure}[t]
    \centering
    \includegraphics[width=1\linewidth]{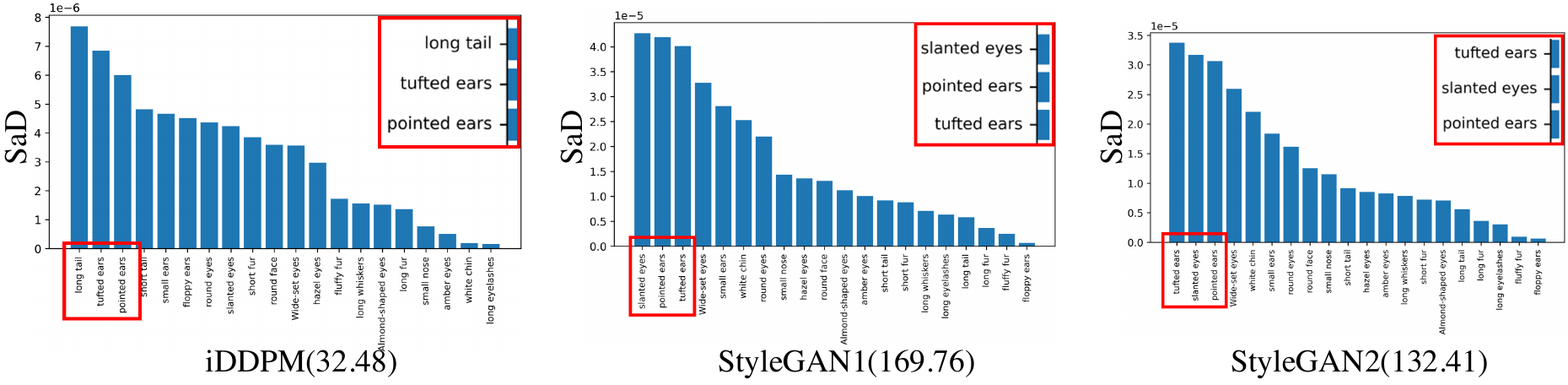}
    \caption{\textbf{SaD for LSUN Cat with shape attributes.}}
    \label{fig:lsun_shape}
\end{figure}

\paragraph{Shape-related attributes}
SaD and PaD of Shape-related attributes were relatively lower than color-related attributes. However, the attributes that have a negative impact on the scores are different in StyleGANs and iDDPM as shown in \fref{fig:lsun_shape}.

Interestingly, among the attributes that DMs struggle with, the worst two attributes, \texttt{long tail} and \texttt{tufted ears}, share the commonality of being thin and long. We speculate that this is similar to the difficulty in creating \texttt{stripes}, indicating a similar characteristic. 

These conjectures also explain why \texttt{arched eyebrows} in FFHQ have a high PaD score. Arched eyebrows have a thin and elongated shape that differs from the typical eyebrow appearance. Considering the characteristics of diffusion models that struggle to create stripes effectively, we can gain insights into the reasons behind this observation.

\begin{table}[t]
\caption{\textbf{Worst 3 \twoabbr{} pair with shape or color attributes on LSUN Cat.}}
\centering
\begin{tabular}{cc|cc|c}
                                                           &     & StyleGAN1                                                           & StyleGAN2                                                            & iDDPM                                                                 \\ \hline
                                                           & 1st & \begin{tabular}[c]{@{}c@{}}fawn fur\\ \&navy fur\end{tabular}  & \begin{tabular}[c]{@{}c@{}}fawn fur\\ \&calcico fur\end{tabular}   & \begin{tabular}[c]{@{}c@{}}tabby fur\\ \&striped fur\end{tabular} \\
\begin{tabular}[c]{@{}c@{}}color\\ attributes\end{tabular} & 2nd & \begin{tabular}[c]{@{}c@{}}fawn fur\\ \&calcico fur\end{tabular} & \begin{tabular}[c]{@{}c@{}}fawn fur\\ \&lilac fur\end{tabular}  & \begin{tabular}[c]{@{}c@{}}dotted fur\\ \&striped fur\end{tabular}     \\
                                                           & 3rd & \begin{tabular}[c]{@{}c@{}}lilac fur\\ \&fawn fur\end{tabular} & \begin{tabular}[c]{@{}c@{}}lilac fur\\ \&navy fur\end{tabular}  & \begin{tabular}[c]{@{}c@{}}black fur\\ \&striped fur\end{tabular}    \\ \hline
                                                           & 1st & \begin{tabular}[c]{@{}c@{}}tufted ears\\ \&slanted eyes\end{tabular}  & \begin{tabular}[c]{@{}c@{}}tufted ears\\ \&slanted ears\end{tabular}  & \begin{tabular}[c]{@{}c@{}}hazel eyes\\ \&long tail\end{tabular}      \\
\begin{tabular}[c]{@{}c@{}}shape\\ attributes\end{tabular} & 2nd & \begin{tabular}[c]{@{}c@{}}pointed ears\\ \&slanted eyes\end{tabular} & \begin{tabular}[c]{@{}c@{}}tufted ears\\ \&white chin\end{tabular}   & \begin{tabular}[c]{@{}c@{}}Almond-shaped eyes\\ \&long tail\end{tabular}    \\
                                                           & 3rd & \begin{tabular}[c]{@{}c@{}}slanted eyes\\ small ears\end{tabular}  & \begin{tabular}[c]{@{}c@{}}pointed ears\\ \&white chin\end{tabular} & \begin{tabular}[c]{@{}c@{}}long tail\\ \&wide-set eyes\end{tabular}     
\end{tabular}
\label{tab:lsun_pad}
\end{table}

\section{Overfitting}
We conducted an experiment to examine a model that performs exact copies of the training set and achieves the highest scores in SaD and PaD.
We created two subsets of real images, Set A and Set B, each containing 30,000 images. Subsequently, we gradually replaced images in Set B with images from Set A, and we measured SaD, PaD and other metrics to understand the impact.
This approach underscores the limitation that an overfitted model may achieve the best score, a challenge inherent to all evaluation metrics, including FID.

Specifically, the tendency due to overfitting reveals that among all metrics, FID, SaD, and PaD all exhibit R-squared values of over 99.5\%, demonstrating a similar level of linearity. These findings indicate that the overfitting tendency prompted by our fine-grained attribute does not worsen.

\begin{table}[ht]
\centering
\caption{\textbf{Performance of various metrics with different levels of image replacement.}}
\begin{tabular}{ccccccccc}
\toprule
\textbf{Number of replaced images} & \textbf{SaD} & \textbf{PaD} & \textbf{FID} & \textbf{FID\_CLIP} & \textbf{Precision} & \textbf{Recall} & \textbf{Density} & \textbf{Coverage} \\
\midrule
100\% & 0.00 & 0.00 & 0.00 & 0.00 & 1.00 & 1.00 & 1.00 & 1.00 \\
90\%  & 0.05 & 0.36 & 0.11 & 0.01 & 0.98 & 0.98 & 0.99 & 0.99 \\
80\%  & 0.12 & 0.77 & 0.21 & 0.02 & 0.97 & 0.98 & 0.99 & 0.99 \\
70\%  & 0.19 & 1.19 & 0.32 & 0.03 & 0.96 & 0.96 & 0.99 & 0.99 \\
60\%  & 0.24 & 1.55 & 0.43 & 0.04 & 0.95 & 0.95 & 0.99 & 0.98 \\
50\%  & 0.29 & 1.91 & 0.54 & 0.05 & 0.95 & 0.95 & 0.99 & 0.98 \\
40\%  & 0.35 & 2.26 & 0.64 & 0.07 & 0.94 & 0.94 & 0.99 & 0.98 \\
30\%  & 0.42 & 2.67 & 0.74 & 0.08 & 0.92 & 0.93 & 0.99 & 0.97 \\
20\%  & 0.46 & 3.04 & 0.85 & 0.09 & 0.90 & 0.90 & 0.99 & 0.97 \\
10\%  & 0.55 & 3.44 & 0.95 & 0.10 & 0.90 & 0.90 & 0.99 & 0.97 \\
0\%   & 0.61 & 3.82 & 1.07 & 0.11 & 0.89 & 0.89 & 0.99 & 0.97 \\
\midrule
\textbf{r\textsuperscript{2} (\%)} & 99.76 & 99.97 & 99.98 & 99.52 & 97.5 & 96.96 & 25.0 & 90.75 \\
\bottomrule
\end{tabular}
\label{tab:performance_metrics}
\end{table}

\section{N-way relationships statistical stability}
Our metric is capable of examining relationships involving any number of attributes (N-way relationships), as it fundamentally relies on measuring the joint probability among involved attributes. The methodology proposed in our paper remains applicable for considering the joint probability of three or more attributes. This is particularly true for exceeding 30k images, where we did not observe statistical instability.

We provide an experiment, structured similarly to \fref{fig:numbers} (metric value's variations across different seeds), to ascertain the minimum number of images that might lead to statistical instability. The proposed metric with attribute triplets demonstrates comparable standard deviation changes with an increased number of images, in comparison to SaD and PaD, indicating statistical stability. Particularly, when measuring with more than 40k images, it exhibits stability with a standard deviation of around 0.1. We conclude that extending our proposed metric to N metrics is feasible, provided that a sufficient number of images are used.

\begin{table}[h]
\centering
\caption{\textbf{Mean and standard deviation of SaD, PaD, and proposed metric with attribute triplets.}}
\resizebox{\textwidth}{!}{
\begin{tabular}{lccccccc}
\toprule
 & 10K & 20K & 30K & 40K & 50K & 60K & 70K \\
\midrule
\textbf{SaD} & 8.56 (0.35) & 8.15 (0.19) & 7.89 (0.19) & 7.96 (0.13) & 7.85 (0.16) & 7.87 (0.11) & 7.83 (0.12) \\
\textbf{PaD} & 24.02 (0.58) & 21.48 (0.38) & 20.32 (0.30) & 20.15 (0.20) & 19.76 (0.25) & 19.64 (0.20) & 19.43 (0.21) \\
\textbf{Attribute Triplets} & 18.67 (0.68) & 16.21 (0.45) & 15.29 (0.46) & 14.31 (0.13) & 13.93 (0.17) & 13.67 (0.08) & 13.44 (0.09) \\
\bottomrule
\end{tabular}}
\end{table}

\begin{table}[h]
\centering
\caption{\textbf{Results for StyleGAN2 and StyleGAN3 trained on AFHQv2.}}
\begin{tabular}{lcc}
\toprule
 & \textbf{SaD} & \textbf{PaD} \\
\midrule
\textbf{StyleGAN2} & 285.09 & 421.90 \\
\textbf{StyleGAN3} & \textbf{281.18} & \textbf{413.53} \\
\bottomrule
\end{tabular}
\end{table}

\section{Evaluating inherent biases in CLIP-like models}

In this appendix, we present the detailed evaluation of inherent biases within various CLIP-like models. The Single-attribute Divergence (SaD) and Paired-attribute Divergence (PaD) metrics were used to assess the biases in ProjectedGAN with a variety of publicly available models similar to CLIP. The results indicated that certain attributes consistently showed high divergence across different models, highlighting underlying biases.


\begin{table}[h]
\centering
\caption{\textbf{Comparison of different models and metrics.}}
\resizebox{\textwidth}{!}{
\begin{tabular}{ll|ccccccc}
\toprule

                                                             &     & \multicolumn{1}{l}{StyleGAN1} & \multicolumn{1}{l}{StyleGAN2} & \multicolumn{1}{l}{StyleGAN3} & \multicolumn{1}{l}{iDDPM} & \multicolumn{1}{l}{LDM (200)} & \multicolumn{1}{l}{StyleSwin} & \multicolumn{1}{l}{ProjectedGAN} \\ \hline
\multicolumn{1}{l|}{\multirow{2}{*}{\textbf{SigLIP}}}        & SaD & 9.31                          & 6.73                          & 6.59                          & 19.62                     & 20.47                         & \textbf{6.41}                 & 13.59                            \\
\multicolumn{1}{l|}{}                                        & PaD & 23.34                         & 17.16                         & \textbf{16.02}                & 43.63                     & 43.62                         & 17.26                         & 33.61                            \\ \hline
\multicolumn{1}{l|}{\multirow{2}{*}{\textbf{CLIP-ViT}}}      & SaD & 11.35                         & \textbf{7.52}                 & 7.79                          & 14.78                     & 14.04                         & 10.76                         & 17.61                            \\
\multicolumn{1}{l|}{}                                        & PaD & 27.25                         & \textbf{19.22}                & 19.73                         & 34.04                     & 30.71                         & 26.56                         & 41.53                            \\ \hline
\multicolumn{1}{l|}{\multirow{2}{*}{\textbf{CLIP-ConvNext}}} & SaD & 17.75                         & 22.09                         & 23.07                         & 26.16                     & 22.34                         & \textbf{9.51}                 & 26.14                            \\
\multicolumn{1}{l|}{}                                        & PaD & 40.22                         & 47.79                         & 49.81                         & 58.45                     & 51.50                         & \textbf{24.16}                & 61.51     \\      
\bottomrule

\end{tabular}}
\end{table}

\begin{table}[h!]
\centering
\caption{\textbf{SaD and PaD values along with the three worst-performing attributes for various CLIP-like models.} We used ProjectedGAN for this experiment.}
\begin{tabular}{l|c|c|cc|c}
\toprule

\multicolumn{1}{c|}{Architecture} & Model    & Dataset      & SaD   & PaD   & Worst 3 attributes                                  \\ \hline
\multirow{2}{*}{\textbf{SigLIP}}           & ViT-B-16 & WebLi        & 13.60 & 33.61 & red lip, child, woman                         \\
                                  & ViT-L-16 & WebLi        & 17.49 & 41.15 & wearing necklace, wearing eyeglasses, woman   \\ \hline
\multirow{7}{*}{\textbf{CLIP-ViT}}         & ViT-B-32 & OpenAI's WiT & 17.61 & 41.53 & makeup, woman, child                          \\
                                  & ViT-B-16 & LAION-2B     & 15.02 & 36.40 & child, wearing necklace, wearing eyeglasses   \\
                                  & ViT-L-14 & LAION-2B     & 23.55 & 54.34 & beard, red lip, woman                         \\
                                  & ViT-H-14 & LAION-2B     & 17.68 & 43.15 & red lip, gray hair, beard                     \\
                                  & ViT-L-14 & DataComp.XL  & 36.30 & 80.02 & woman, child, blonde hair                     \\
                                  & ViT-L-14 & DFN-2B       & 16.32 & 40.17 & child, smiling, double chin                   \\
                                  & ViT-H-14 & DFN-5B       & 21.24 & 52.03 & wearing eyeglasses, woman, wearing necklace   \\ \hline
\multirow{2}{*}{\textbf{CLIP-ConvNext}}    & base\_w   & LAION-2B     & 13.60 & 33.61 & bald head, wearing necklace, wearing earrings \\
                                  & large\_d  & LAION-2B     & 17.49 & 41.15 & makeup, smiling, young          \\              
\bottomrule

\end{tabular}
\end{table}

\paragraph{Impact of model variation on SaD and PaD.}

To further understand the impact of model variation, we analyzed the SaD and PaD metrics across different generative models. Despite some differences in the SaD and PaD values, the overall trend remained consistent, indicating similar biases across the models.

\paragraph{Consistent trends and future research directions}

Despite the variability in the values introduced by different CLIP-like models, certain attributes such as \texttt{makeup}, \texttt{woman}, \texttt{red lip}, \texttt{wearing necklace}, \texttt{child}, and \texttt{young} consistently showed high divergence. This indicates a uniform trend of bias across models. Addressing the variations in result values caused by changes in the encoder presents an interesting avenue for future research. Notably, StyleSwin demonstrated strong performance with certain CLIP models, suggesting potential pathways for mitigating these biases.

\end{appendix}




\end{document}